\documentclass[10pt,twocolumn,letterpaper]{article}

\usepackage[pagenumbers]{cvpr} 
\usepackage{times}
\usepackage{epsfig}
\usepackage{graphicx}
\usepackage{amsmath}
\usepackage{amssymb}
\usepackage{booktabs}
\usepackage{multirow}
\usepackage{tabu}
\usepackage{bm}
\usepackage{array}
\usepackage{fontawesome}
\usepackage{verbatim}
\usepackage{makecell} 
\usepackage{algorithm}
\usepackage{listings}
\usepackage[section]{placeins}
\usepackage[usenames,dvipsnames]{color}
\usepackage{colortbl}
\usepackage[table,dvipsnames]{xcolor}
\usepackage[accsupp]{axessibility}
\usepackage{subcaption}

\usepackage[pagebackref=true,breaklinks=true,letterpaper=true,colorlinks,bookmarks=false]{hyperref}

\usepackage[capitalize]{cleveref}
\crefname{section}{Sec.}{Secs.}
\Crefname{section}{Section}{Sections}
\Crefname{table}{Table}{Tables}
\crefname{table}{Tab.}{Tabs.}

\definecolor{mygray}{gray}{.9}
\definecolor{LightCyan}{rgb}{0.88,1,1}
\definecolor{lightgray}{RGB}{128,128,128}
\definecolor{lightblue}{RGB}{212,239,251}
\definecolor{lightgray2}{RGB}{235,235,235}

\begin{document}

\title{HaMuCo: Hand Pose Estimation via Multiview Collaborative \\Self-Supervised Learning}

\author {
    Xiaozheng Zheng \textsuperscript{\rm 1,2} \qquad
    Chao Wen \textsuperscript{\rm 2} \qquad
    Zhou Xue \textsuperscript{\rm 2} \qquad
    Pengfei Ren \textsuperscript{\rm 1,2} \qquad
    Jingyu Wang \textsuperscript{\rm 1 \footnotemark[1]} \\
    \textsuperscript{\rm 1} State Key Laboratory of Networking and Switching Technology, BUPT \\
    \textsuperscript{\rm 2} PICO IDL, ByteDance, Beijing \\
}

\maketitle

{
  \renewcommand{\thefootnote}%
    {\fnsymbol{footnote}}
  \footnotetext{Project page: \url{https://zxz267.github.io/HaMuCo}.}
}

{
  \renewcommand{\thefootnote}%
    {\fnsymbol{footnote}}
  \footnotetext[1]{Corresponding author.}
}

\begin{abstract}
Recent advancements in 3D hand pose estimation have shown promising results, but its effectiveness has primarily relied on the availability of large-scale annotated datasets, the creation of which is a laborious and costly process. To alleviate the label-hungry limitation, we propose a self-supervised learning framework, HaMuCo, that learns a single-view hand pose estimator from multi-view pseudo 2D labels. However, one of the main challenges of self-supervised learning is the presence of noisy labels and the ``groupthink'' effect from multiple views. To overcome these issues, we introduce a cross-view interaction network that distills the single-view estimator by utilizing the cross-view correlated features and enforcing multi-view consistency to achieve collaborative learning. Both the single-view estimator and the cross-view interaction network are trained jointly in an end-to-end manner. Extensive experiments show that our method can achieve state-of-the-art performance on multi-view self-supervised hand pose estimation. Furthermore, the proposed cross-view interaction network can also be applied to hand pose estimation from multi-view input and outperforms previous methods under the same settings.
\end{abstract}

\vspace{-0.3cm}
\section{Introduction}

\begin{figure*}[t]
        \vspace{-0.4cm}
	\centering
	\includegraphics[width=\textwidth]{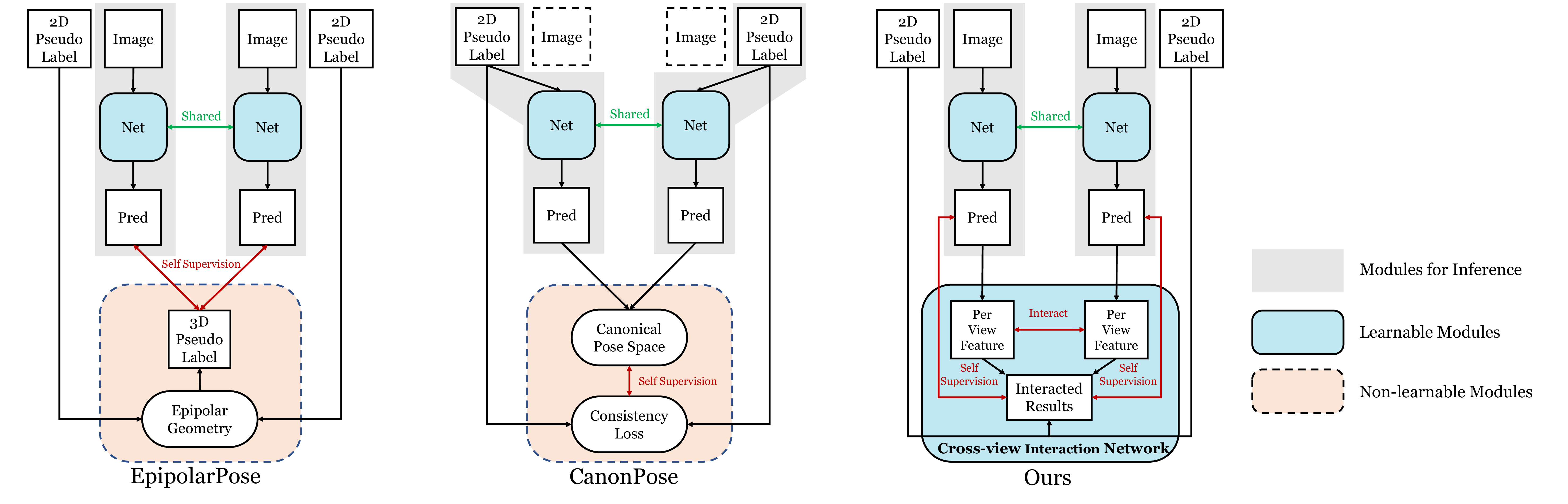}
        \vspace{-2.2em}
	\caption{Overall pipeline comparison: HaMuCo learns a monocular 3D hand pose estimator from multi-view self-supervision via cross-view feature interaction. Our cross-view interaction network addresses the importance of introducing learnable feature interaction, which is absent from previous methods~\cite{wandt2021canonpose,kocabas2019self}. At inference time only the gray part is applied.}
	\label{img:iccv-teaser}
    \vspace{-1.5em}
\end{figure*}

\label{sec:intro}
3D hand pose estimation is essential in various application scenarios, from action recognition and sign language translation to AR/VR~\cite{han2020megatrack,han2022umetrack}. 
Hand pose estimation has achieved a significant improvement in recent years. However, the progress heavily relies on the emergence of many hand pose datasets with accurate 3D annotations. Acquiring labeled datasets is quite time-consuming and laborious, exposing a realistic challenge for deep learning models to learn with limited and noisy data. 

Self-supervised learning is an emerging solution to the challenge posed by manual annotation. 
Though worth exploring, self-supervised pose estimation with RGB hand images is a challenging and relatively unexplored area with only one pioneering method, S$^2$HAND~\cite{chen2021model}.
S$^2$HAND aims to conduct 3D hand reconstruction from a single RGB image with the noisy off-the-shell 2D hand pose estimation results~(OpenPose) for supervision. Unfortunately, S$^2$HAND faces a predicament where its performance is significantly reliant on the quality of the pseudo label, and inferior labeling may result in reduced performance. Moreover, evaluating the quality of the pseudo label is an ill-posed problem that lacks clear criteria or input, further complicating the issue.

This observation motivates us to leverage multi-view information for enhancing self-supervised learning, as the complementary nature of multi-view observations can help mitigate the ambiguity inherent in pose estimation. Although the first 3D hand dataset with synchronized multi-view input (HanCo~\cite{zimmermann2021contrastive}) was proposed in 2021, to our knowledge, there is no previous work exploring the potential of multi-view for self-supervised hand pose estimation. Therefore, we turn to studies in the human body pose estimation, which share some similarities. 

As mentioned in previous work~\cite{iqbal2020weakly}, naively enforcing multi-view consistency is prone to generate degenerated solutions, thus they resorted to additional 2D labels of unrelated datasets and proposed a solution under the scope of weakly supervised learning. Other studies, such as EpipolarPose\cite{kocabas2019self} and CanonPose \cite{wandt2021canonpose}, utilized multi-view data with special designs to enhance the supervision and achieved promising results under the scope of self-supervised learning.

In this paper, we push along this direction on hand pose estimation via multi-view collaborative learning. We take one step further by designing a learnable network, which utilizes multi-view information, to tackle 1) noisy pseudo labels and 2) unreliable multi-view ``groupthink'' issues causing training collapse in the early training stage. Formally, we name the pipeline HaMuCo, which stands for \underline{Ha}nd \underline{Mu}ltiview \underline{Co}llaborative learning.

The core idea of our approach is to enhance the single-view estimation by means of cross-view feature interaction and further integrate multi-view results to supervise the single-view output to achieve self-distillation in an end-to-end fashion. Thus, our framework is built with a single-view hand pose estimator and a cross-view interaction network for supervision. The single-view estimator uses the MANO~\cite{romero2017embodied} hand model as the decoder, which provides the hand prior to regularizing irrational anatomy when supervised by noisy pseudo labels. The cross-view interaction network captures cross-view features and utilizes several consistent losses among different views to guide collaborative learning.

We conduct comprehensive experiments on the HanCo~\cite{zimmermann2021contrastive} dataset and our approach outperforms previous methods by a considerable margin for self-supervised 3D hand pose estimation. Notably, our results demonstrate competitive performance compared to a state-of-the-art fully supervised approach proposed by Chen \etal~\cite{chen2022mobrecon}. Our proposed framework is highly versatile, as it can be trained with or without calibration, and is capable of incorporating the cross-view interaction network to achieve superior multi-view inference results when multi-view test data is available. Moreover, we show that our model can generalize well to other datasets \cite{zimmermann2019freihand,kwon2021h2o,sener2022assembly101} and in-the-wild images.

In summary, our contributions are the following: 
 \begin{itemize}
     \item We propose the first self-supervised learning framework for single-view hand pose estimation without any training data annotation and achieve state-of-the-art performance by via multi-view collaborative learning.
     \item We propose a cross-view interaction network to supervise the single-view estimator by enforcing multi-view consistency and capturing cross-view features for collaborative learning among multiple views.
     \item The proposed framework is capable of multi-view inference by incorporating the cross-view interaction network and achieves state-of-the-art performance without bells and whistles. 
 \end{itemize}

\section{Related Work}
\label{sec:related work}
\vspace{-0.3em}
\noindent
\textbf{Hand Pose Estimation.}
Hand pose estimation can be categorized into RGB-based methods \cite{zimmermann2017learning,spurr2018cross,iqbal2018hand} and depth-based methods \cite{ge2016robust,ge20173d,moon2018v2v}, depending on the input modality. In this paper, we focus our attention on RGB-based hand pose estimation. The RGB-based methods can be further divided into three categories, \emph{skeleton-based methods} \cite{zimmermann2017learning,spurr2018cross,iqbal2018hand,cai2018weakly,mueller2018ganerated,yang2019aligning,yang2019disentangling,spurr2020weakly,doosti2020hope,moon2020interhand2,li2021exploiting,yang2021semihand}, \emph{model-based methods} \cite{boukhayma20193d,baek2019pushing,zhang2019end,zimmermann2019freihand,baek2020weakly,chen2021model,zhang2021hand}, and \emph{mesh-based} methods\cite{ge20193d,kulon2020weakly,moon2020i2l,choi2020pose2mesh,chen2021camera,zheng2021sar,lin2021end,lin2021mesh,tang2021towards,chen2022mobrecon,li2022interacting}. 
\emph{Skeleton-based methods} regress the hand joints directly. 
Zimmermann \etal \cite{zimmermann2017learning} introduces a multi-stage network that lifts the regressed 2D joints to 3D ones.
Variational autoencoder \cite{kingma2013auto} is employed to learn a cross-modal latent space to achieve better hand pose estimation and disentanglement \cite{spurr2018cross,yang2019disentangling,yang2019aligning}. 
Latent 2.5D representation regression is proved more effective than direct coordinates regression for hands by \cite{iqbal2018hand}, which is also adopted by \cite{spurr2020weakly,li2021exploiting,zheng2021sar,fan2021learning}. 
There are also many works solving hand pose estimation with two hands interactions \cite{moon2020interhand2,fan2021learning,kwon2021h2o} and hand-object interactions \cite{doosti2020hope,kwon2021h2o,baek2020weakly}.
Recent \emph{model-based methods} make use of MANO \cite{romero2017embodied}, which can incorporate the hand prior and predict the hand mesh simultaneously. Those methods \cite{boukhayma20193d,baek2019pushing,zhang2019end,chen2021model,zhang2021hand} rely on additional supervisions \cite{boukhayma20193d,baek2019pushing,zhang2019end,chen2021model,zhang2021hand} or inputs \cite{boukhayma20193d}. In contrast, \emph{mesh-based methods} regress each vertex directly, which is more accurate but requires large-scale datasets with hand mesh annotations \cite{zimmermann2019freihand,hampali2020honnotate,moon2020interhand2,kwon2021h2o}. Most of these methods utilize graph convolutional network~(GCN) \cite{ge20193d,kulon2020weakly,choi2020pose2mesh,chen2021camera,tang2021towards,zheng2021sar,chen2022mobrecon} or transformers\cite{lin2021end} or both \cite{lin2021mesh,li2022interacting} for regression. I2L-MeshNet \cite{moon2020i2l} regresses each vertex by predicting 1D heatmaps of three axes. Chen \etal \cite{chen2021i2uv} uses an image-to-image translation network to predict the UV map of the mesh. 
Similar to previous works \cite{lin2021mesh,li2022interacting}, we also use transformer and GCN. However, we employ them for cross-view interaction. 

\noindent
\textbf{Multi-View Fully-Supervised Pose Estimation.}
Multi-view information is widely explored to improve 3D human pose estimation by tackling occlusions and depth-ambiguity in a fully-supervised manner\cite{pavlakos2017harvesting,bartol2022generalizable,iskakov2019learnable,qiu2019cross,he2020epipolar,remelli2020lightweight,zhang2021adafuse,shuai2022adaptive}. Volume-based methods \cite{pavlakos2017harvesting,iskakov2019learnable,qiu2019cross,tu2020voxelpose} unproject 2D features or heatmaps of joints to a 3D space for estimation. 
Another kind of method \cite{he2020epipolar,zhang2021adafuse,remelli2020lightweight} utilizes the geometry information to fuse the features in 2D space directly and efficiently. 
Recently, some works \cite{shuai2022adaptive,ma2022ppt} utilize transformers for implicit cross-view fusion without camera extrinsics.

\noindent
\textbf{Label-Efficient Learning.}
Label-efficient learning aims to reduce the 3D label requirements. Many works devote to solving hand pose estimation in a label-efficient manner \cite{zimmermann2017learning, mueller2018ganerated,cai2018weakly,boukhayma20193d,baek2019pushing,zhang2019end,spurr2020weakly,yang2021semihand,chen2021model,chen2022mobrecon}. 
Synthetic data is used to avoid manual annotation~\cite{zimmermann2017learning,mueller2018ganerated,chen2022mobrecon}, but may need domain transfer \cite{mueller2018ganerated}. 
Or use weakly supervised learning~\cite{boukhayma20193d,spurr2020weakly} to obtain 3D results by manually annotating 2D labels to assist with hand priors.
Multi-view label-efficient learning is also explored in 3D pose estimation \cite{rhodin2018learning,kocabas2019self,iqbal2020weakly,wandt2021canonpose}.
Rhodin \etal \cite{rhodin2018learning} trains a semi-supervised network with only a small amount of labeled 3D data and multi-view consistency constraints. Iqbal \etal \cite{iqbal2020weakly} mixes single-view images with 2D labels and unlabelled multi-view images for training. 
Our goal is the same as that of previous methods, which is to train without any manual 3D labels.

\noindent
\textbf{Self-supervised 3D Pose Estimation.}
(1) Single-view training and inference.
To the best of our knowledge, there is only one method for self-supervised 3D hand pose estimation, proposed by Chen \etal \cite{chen2021model}. Their framework, S$^2$Hand, uses only single-view 2D noisy labels for training and achieves self-supervision through rendering. However, the performance is limited due to the use of single-view information and the quality of the noisy labels.
(2) Multi-view training, single-view inference.
Our approach belongs to this category but is fundamentally different from the existing methods. 
EpipolarPose~\cite{kocabas2019self} triangulates multi-view 2D pseudo labels according to epipolar geometry to 3D ones for training.
CanonPose~\cite{wandt2021canonpose} learns to lift 2D pseudo labels to 3D canonical pose space with multi-view consistency constraints.
All the aforementioned methods use non-learnable self-supervised modules like geometric modules or consistency loss functions, as shown in \cref{img:iccv-teaser}.
However, they~\cite{wandt2021canonpose,kocabas2019self} ignore the importance of introducing cross-view interaction and multi-view collaborative learning.
Previous methods struggle to achieve good performance since the pose of a hand can change drastically over time and different joints may have similar appearances.

\section{Method}

\begin{figure*}[!ht]
	\centering
	\includegraphics[width=\textwidth]{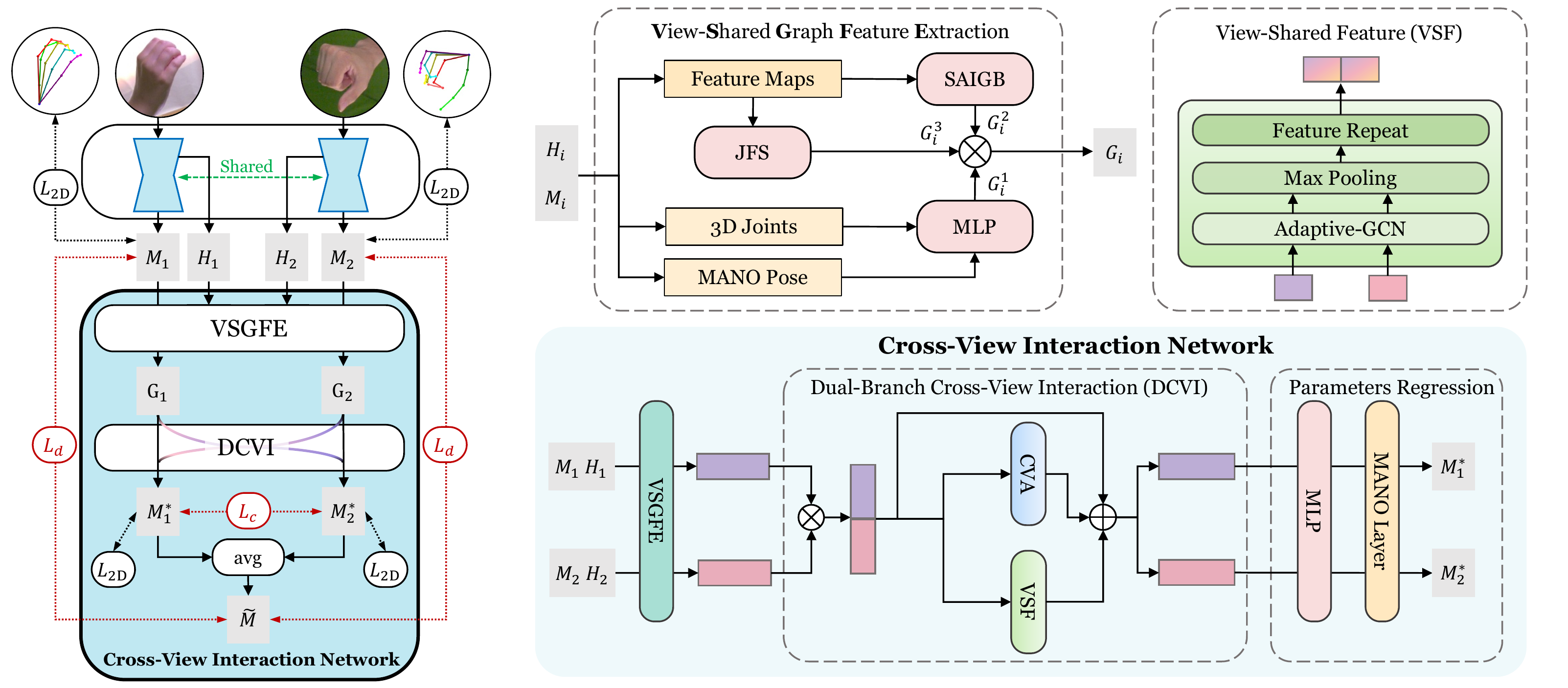}
 	\caption{The left illustrates our whole pipeline (2 views for simplicity). During the training phase, the network takes multi-view hand images and pseudo-labels as inputs. The bottom right depicts our cross-view interaction networks. The top right shows the view-shared graph feature extraction (VSGFE) module and view-shared feature (VSF) module. $\oplus$ and $ \otimes$ denotes add and concatenation respectively.}
	\label{fig:pipeline} 
	\vspace{-1em}
\end{figure*}

\label{sec:method}
As depicted in the left part of~\cref{fig:pipeline}, our framework consists of a simple yet effective single-view estimator and cross-view interaction network. 
The core idea of our approach is that prediction from a monocular view can be enhanced via cross-view feature interaction and the interacted results can further supervise the single-view output to achieve self-distillation.

\subsection{Single-View Estimator}\label{sec:sve}
\noindent
\textbf{Overview.}
Our framework takes multi-view synchronized hand images $\mathcal{I} = \{\bm{I}_{i}\}_{i=0}^{v}$ with $v$ views as input, each view is an image of $\bm{I}_{i} \in \mathbb{R}^{3\times h\times w}$.
The output is a 3D hand mesh $\bm{M}$ on each view.
We designed a simple yet effective model-based network as a single-view estimator.
Using the hand model will reduce the adverse effects of using poor pseudo labels as supervision by providing hand prior information for regularization.
Please refer to supplementary materials for more details about the single-view estimator.

\noindent
\textbf{Hand model.} We employ MANO~\cite{romero2017embodied} as the hand model.
The hand mesh can be derived from the MANO layer using parameters $\bm{\beta}$ and $\bm{\theta}$, \ie $\bm{M}(\bm{\beta}, \bm{\theta})$.
$\bm{\beta} \in \mathbb{R}^{10}$ and $\bm{\theta} \in \mathbb{R}^{16 \times 3}$ control the shape and pose of the hand respectively.
We can use a predefined regressor to obtain the 3D joints from the 3D mesh vertices by $\bm{P} = \bm{J}\bm{M}$, where $\bm{J} \in \mathbb{R}^{k\times n}$, where $n=778$ and $k=21$ are the joints number and vertices number. For more details, we recommend referring to \cite{romero2017embodied}.

\noindent
\textbf{Camera model.} 
Following Boukhayma~\etal~\cite{boukhayma20193d}, we model the geometry correspondence by the weak-perspective camera model and obtain camera parameters from the single-view network predictions.
Given the translation $\bm{t}$ and scale $s$, the 2D coordinates in image plane can be obtained by: $\Pi(\bm{P})=s\Omega(\bm{P})+\textbf{t}$, where $\Omega$ is the orthographic projection and $\Pi$ denotes the weak-perspective projection.

\noindent
\textbf{Network Structure.}
Since the single-view estimator is not the main component, for the sake of simplicity, we employ a CNN as the encoder $F_{e}$, and an MLP as the decoder $F_{d}$ for regressing the MANO parameters.
We have 3D hand mesh: $\bm{M}_{i}(\bm{\theta}_{i}, \bm{\beta}_{i}) = F_{s}\left(\bm{I}_{i}\right)$, where $F_{s}= F_{d}\left(F_{e}\left(\cdot \right)\right)$ denotes the entire single-view network. 
The estimator also passes different levels of features $\bm{H}^{j}$~(where $\bm{H}^{j}$ is the intermediate feature of the encoder after $j$ residual blocks, $j$=$1,2,3,4$) to our cross-view interaction network.

\subsection{Cross-view Interaction Network}\label{sec:cvi}
In this section, we introduce the cross-view interaction network (CVI-Net), which is the core of our system to enable the network to exploit multi-view information.
This stage conducts cross-view interaction and distillation.
The critical components of this stage are a cross-view interaction network for capturing cross-view features and several consistent losses for guiding collaborative learning.

\vspace{-1em}
\subsubsection{View-Shared Graph Feature Extraction}
\vspace{-0.5em}
The first step for interaction is to extract the appropriate features. 
Different from~\cite{chen2021camera,zheng2021sar,chen2022mobrecon}, our module collects useful information into a graph through view-shared graph feature extraction module~(VSGFE) as shown in \cref{fig:pipeline}.
Specifically, it makes use of multi-level feature maps from different views $\mathcal{H} = \{  \bm{H}_{i}^{j}\}_{i=0}^{v}$, 3D joints $\mathcal{P} = \{\bm{P}_{i}\}_{i=0}^{v}$, and MANO pose parameters $\Theta = \{\bm{\theta}_{i}\}_{i=0}^{v}$ from the single-view estimator to extract a graph feature $\bm{G}$.
The graph feature of each view $\bm{G}_{i}$ consists of three parts.
$\bm{G}_{i}^{1}$, $\bm{G}_{i}^{2}$ and $\bm{G}_{i}^{3}$ aim to capture joint location features, global image features, and local image features, respectively.
The first part is joint location embedding $\bm{G}_{i}^{1} \in \mathbb{R}^{k \times c_{1}}$, providing the explicit geometric information.
This embedding is obtained by using an MLP to map the single-view 3D joints locations $\bm{P}_{i}$ and pose parameters $\bm{\theta}_{i}$ to dimension $c_{1}$.
The second part is joint-wise high-level image features $\bm{G}_{i}^{2} \in \mathbb{R}^{k \times c_{2}}$ generated by spatial-aware initial graph building~(SAIGB)~\cite{zheng2021sar} module using the last level feature maps $ \bm{H}_{i}^{4}$. This part provides compact image clues of all views for interaction.
The third part is joint-aligned features $\bm{G}_{i}^{3} \in \mathbb{R}^{k \times c_{3}}$ gathered by joint feature sampler (JFS). 
JFS projects joints onto multi-level image feature maps $\{  \bm{H}_{i}^{j}\}_{j=1}^{3}$ to gather fine-grained perceptual features like~\cite{wang2018pixel2mesh,wen2019pixel2mesh++} for better local alignment.
We then concatenate graph features to get $\bm{G}_{i} = \left [\bm{G}_{i}^{1} \otimes \bm{G}_{i}^{2} \otimes  \bm{G}_{i}^{3} \right ]$. 

\vspace{-1.5em}
\subsubsection{Dual-Branch Cross-View Interaction (DCVI)}
\vspace{-0.5em}
We first stack $\{\bm{G}_{i}\}_{i=0}^{v}$ of all views to obtain multi-view graph feature $\bm{G} \in \mathbb{R}^{vk \times (c_{1}+c_{2} +c_{3})}$.
We design a component to effectively capture complementary information from other views on multi-view graph feature $\bm{G}$.
The interaction module has two branches, (1)\;\emph{cross-view attention branch}\;(CVA) and (2)\;\emph{view-shared feature branch}\;(VSF).
\emph{Cross-view attention branch} utilizes a cross-view transformer $F_{t}$ consisting of several multi-head attention layers with token size $vk$ and MLPs, which allows each joint to aggregate features from other joints or views.
This branch implicitly captures the multi-view information.
An explicit multi-view prior information is that the observed poses from all the views should be consistent in 3D. Therefore, we add a branch to excavate the multi-view shared information to enhance the feature representation.
Specifically, \emph{view-shared feature branch} first employs adaptive-GCN \cite{doosti2020hope} $F_{a}$ to map the view-specific features $\bm{G}_{i}$ to a canonical feature space $\bm{C}_{i}=F_{a}(\bm{G}_{i})$, the nodes in adaptive-GCN represents the hand joints and the edges represents joint feature correlation.
Then, we stack $\mathcal{C} = \{\bm{C}_{i}\}_{i=0}^{v}$ together to get multi-view canonical features $\bm{C} \in \mathbb{R}^{v \times k \times (c_{1}+c_{2}+c_{3})}$. After that, we use max-pooling on $\bm{C}$ to get the max activated features of every joint then repeat them in the view dimension as the view-shared features $\bm{C}^{'} \in \mathbb{R}^{vk \times (c_{1}+c_{2}+c_{3})}$.
We denote the dual-branch cross-view interaction as:
$\bm{G}^{*} = \bm{G} + F_{t}(\bm{G}) + \bm{C}^{'}$,
where $\bm{G}^{*}$ is the updated graph feature.

\noindent
\textbf{Parameters regression.}
The view specific feature $\bm{G}^{*}_{i}$ after the interaction can be obtained by reshaping $\bm{G}^{*}$. We then employ a shared MLP $F_{r}$ as a decoder to regress the pose parameters $\bm{\theta}^{*}_{i}=F_{r}(\bm{G}^{*}_{i})$ to derive the hand mesh of each view $\bm{M}^{*}_{i}(\bm{\theta}^{*}_{i}, \bm{\beta}_{i})$ and corresponding joints $\bm{P}^{*}_{i} = \bm{J}\bm{M}^{*}_{i}$.

\vspace{-1em}
\subsubsection{Multi-View Collaborative Learning}\label{sec:mvc}
To allow all the views and the networks to learn collaboratively, we utilize consistency losses $L_{c}$ upon interaction outputs and distillation loss $L_{d}$ between multi-view fusion results and single-view outputs, as shown in \cref{fig:pipeline}.
$L_{c}$ introduces collaborative learning between multiple views, guiding the poses from different views to be as close as possible.
While $L_{d}$ makes the CVI-Net and single-view estimator work in a collaborative manner, achieving a self-distillation effect.

\noindent
\textbf{Results fusion.} 
Since we need to supervise the single-view estimator with the results after the interaction, instead of simply using the refined results $\bm{M}^{*}_{i}$ of each view, we ensemble all the results into a unified and more reliable result $\tilde{\bm{M}}$. 
Considering the lack of explicit guidance, we empirically introduce a prior that all the views contribute equally. Thus, we simply average all aligned results to obtain $\tilde{\bm{M}}$.
Specifically, we use $A$ to denote the align procedure.
When the extrinsics are known, we use the relative camera pose for alignment. When the camera extrinsics are unavailable, we use Procrustes analysis~\cite{zimmermann2019freihand,zimmermann2021contrastive} to compute relative rotation and align meshes to a canonical view.
The final result is calculated as follows: $\tilde{\bm{M}} = \frac{1}{v}\sum_{i=1}^{v} A(\bm{M}^{*}_{i})$.

\noindent
\textbf{Consistency losses.}
We design two types of consistency loss $L_{c}$: \emph{2D consistency loss} $L_{c_{2D}}$ and \emph{Fusion consistency loss} $L_{c_{f}}$. The motivation behind $L_{c_{2D}}$ is that the 2D predictions in the x-axis and y-axis are more accurate than the depth prediction in the z-axis. Therefore, $L_{c_{2D}}$ utilizes the 2D predictions in every single view as the pseudo label to supervise other views, which explores the view-specific reliable information to collaboratively improve the predictions of all the views.
\emph{2D consistency loss} is defined as: $L_{c_{2D}} = \frac{1}{v^{2}}\sum_{i=1}^{v} \sum_{j=1}^{v} \|\Pi(\bm{M}^{*}_{i}) - \Pi(A_{i}(\bm{M}^{*}_{j}))\|_{1}$, where ${A}_{i}(\cdot)$ denotes the alignment operation to align other view-$j$ to view-$i$. 
\emph{Fusion consistency loss} uses the fused results to supervise each view. The loss is defined as:$L_{c_{f}} = \frac{1}{v} \sum_{i=1}^{v} \|\bm{M}^{*}_{i} - A^{-1}_{i}( \tilde{\bm{M}})\|_{1}$, 
where ${A}^{-1}_{i}(\cdot)$ denotes the inverse transformation from canonical view to view-$i$.
$L_{c_{2D}}$ and $L_{c_{f}}$ are complementary to each other. Only using $L_{c_{2D}}$ tends to get performance saturation faster. In contrast, only adopting $L_{c_{f}}$ can lead to unstable training since there may exist the situation that the fusion results are worse due to the majority of the predictions being wrong, especially at the early training stage. During training, we alternately update $L_{c_{2D}}$ and $L_{c_{f}}$ to achieve more stable optimization. 

\noindent
\textbf{Multi-view distillation loss.}
Since the multi-view fusion results are much better than the 2D pseudo label, we introduce multi-view distillation loss $L_{d} = \frac{1}{v} \sum_{i=1}^{v} \|\bm{M}_{i} - A^{-1}_{i}(\tilde{\bm{M}}) \|_{1}$ that uses the fusion results to supervise the single-view outputs to achieve self-distillation.

\noindent
\textbf{Total loss.} Except for the losses for multi-view collaborative learning, our framework also adopts two general constraints, 2D joints loss, and hand prior regularization. 
The prior regularization regularizes the pose and shape parameters: $L_{p} = \frac{1}{v} \sum_{i=1}^{v} \alpha (\|\bm{\theta}_{i}\|_{1} + \|\bm{\theta}^{*}_{i}\|_{1} + \gamma \|\bm{\beta}_{i}\|_{1})$, where $\alpha$ and $\gamma$ are used to balance the loss scale. The 2D joints $L1$ loss $L_{2D}$ is used to supervise the results from the 2D pseudo labels.
The final loss is defined as: 
\begin{equation}
L = L_{c} + L_{d} + L_{2D} + L_{p}.
\end{equation}

\section{Experiments}
\label{sec:experiments}

\begin{table*}[!ht]
    \hspace{0.5cm}
    \centering
    \makebox[0pt][c]{\parbox{1.05\textwidth}{
        \begin{minipage}[h]{0.35\textwidth}
            \centering
            \begin{minipage}[t]{1.3\textwidth}

                \centering
                \scalebox{0.93}{
                    \begin{tabular}{cccc}
                        \toprule
                        {Method}                            & {Input}                           & {N-JE$\downarrow$} & {PA-JE$\downarrow$} \\
                        \midrule
                        \multicolumn{2}{l}{\emph{\textcolor{lightgray}{\small{Fully-Supervised Method:}}}} \\
                        MobRecon \cite{chen2022mobrecon}    & image                             & 9.9                & 5.7                 \\
                        EpipolarPose \cite{kocabas2019self} & image                             & 10.5               & 6.1                 \\

                        \midrule

                        \multicolumn{2}{l}{\emph{\textcolor{lightgray}{\small{Self-Supervised Method:}}}} \\
                        EpipolarPose \cite{kocabas2019self} & image, \footnotesize{\faCamera}   & 19.7               & 9.3                 \\
                        CanonPose \cite{wandt2021canonpose} & 2D pose, \footnotesize{\faCamera} & 30.9               & 12.6                \\
                        \rowcolor{lightblue}
                        Ours                                & image, \footnotesize{\faCamera}   & \textbf{11.1}      & \textbf{7.0}        \\
                        EpipolarPose \cite{kocabas2019self} & image                             & 42.3               & 23.5                \\
                        CanonPose \cite{wandt2021canonpose} & 2D pose                           & 31.8               & 12.8                \\
                        \rowcolor{lightblue}
                        Ours                                & image                             & \textbf{15.2}      & \textbf{7.7}        \\
                        \bottomrule
                    \end{tabular}}

                \caption{Single-view inference comparisons on the HanCo~\cite{zimmermann2021contrastive} dataset. {\footnotesize{\faCamera}} denotes the method using camera extrinsics during training. Notably, in the self-supervised setting, our method exhibits a significant improvement over previous methods.}
                \label{tab:single-view inference - hanco}
            \end{minipage}
        \end{minipage}
            \hspace{0.7cm}
            \begin{minipage}[h]{0.7\textwidth}
                \centering            
                \vspace{-0.3cm}
                \begin{minipage}[t]{0.75\textwidth}
                    \centering
                    \scalebox{0.85}{
                        \begin{tabular}{ccccccc}
                            \toprule
                            Method                             & Data  & Backbone          & PA-JE$\downarrow$ & PA-VE$\downarrow$ & F@5$\uparrow$ \\
                            \midrule
                            \multicolumn{3}{l}{\emph{\textcolor{lightgray}{\small{{Fully-Supervised Method:}}}}}
                            \\
                            YoutubeHand \cite{kulon2020weakly} & Frei. & Res50             & 8.4               & 8.6               & 0.61          \\
                            I2UV-HandNet \cite{chen2021i2uv}   & Frei. & Res50             & 6.7               & 6.9               & 0.71          \\
                            MobRecon \cite{chen2022mobrecon}   & Frei. & Res50$^{\dagger}$ & 6.1               & 6.2               & 0.76          \\
                            \rowcolor{lightgray2}
                            Ours-SV    & Frei. & Res50 & 7.5               &  7.5        & 0.68          \\
                            \midrule
                            \multicolumn{3}{l}{\emph{\textcolor{lightgray}{\small{Self-Supervised Method:}}}}
                            \\
                            S$^{2}$HAND\cite{chen2021model}    & Frei. & EffiNet-b0        & 11.8              & 11.9              & 0.48          \\
                            \rowcolor{lightgray2}
                            Ours-SV                     & Frei. & EffiNet-b0             & 11.6              & 11.7          & 0.49         \\
                            \rowcolor{lightgray2}
                            Ours-SV                   & Frei. & Res50             & 11.9              & 12.0              & 0.47          \\
                            \rowcolor{lightgray2}
                            Ours-SV                     & HanCo & EffiNet-b0             & 11.3              & 11.4            & 0.51          \\
                            \rowcolor{lightgray2}
                            Ours-SV                     & HanCo & Res50             & 11.6              & 11.8              & 0.48          \\
                            \rowcolor{lightblue}
                            Ours                               & HanCo & EffiNet-b0             & 6.3              & 6.8              & 0.71          \\
                            \rowcolor{lightblue}
                            Ours                               & HanCo & Res50             & \textbf{6.2}               & \textbf{6.7}               & \textbf{0.72}          \\
                            \bottomrule
                        \end{tabular}
                    }
                    \caption{Quantitative results on the FreiHAND evaluation set. The notation $^{\dagger}$ denotes using a stacked backbone structure. "Our-SV" refers to training only with our single-view network. 
                    }
                    \label{tab:single-view inference - freihand}
                \end{minipage}%
            \end{minipage}
            
        }}
    \vspace{-0.4cm}
\end{table*}

\begin{table}[b]
\vspace{-1em}
  \centering
  \scalebox{0.85}{
  \begin{tabular}{lcccccccc}
        \toprule
       Method & MPJPE~$\downarrow$ & PA-MPJPE~$\downarrow$  \\
      \midrule
      \multicolumn{3}{l}{\emph{\textcolor{lightgray}{\small{Ttraditional Triangulation Method (w/o training):}}}} \\
      DLT \cite{hartley2003multiple}       & 16.8 & 13.2 \\
      Pictorial \cite{dong2019fast}     & 13.5 & 10.2  \\
      RANSAC \cite{iskakov2019learnable}    & 12.3 & 9.8 \\
      \midrule
        \multicolumn{3}{l}{\emph{\textcolor{lightgray}{\small{Fully-Supervised Method:}}}} \\
       EpipolarTrans~\cite{he2020epipolar} & 6.2 & 4.2 \\
       LT-Algebraic \cite{iskakov2019learnable} & 5.5 & 3.6 \\
       LT-Volumetric \cite{iskakov2019learnable} & 5.8 & 3.6  \\
       LT-Volumetric$^{+}$ \cite{iskakov2019learnable} & \textbf{4.9} & 3.6 \\
       EpipolarPose$^{+}$ \cite{kocabas2019self}   & 8.0 & 4.4   \\
      \rowcolor{lightblue}
       Ours (Opt-Center)  & 6.0 & \textbf{3.2} \\
       \rowcolor{lightblue}
       Ours (RANSAC)   & 5.8 & 3.4  \\
      \midrule
        \multicolumn{3}{l}{\emph{\textcolor{lightgray}{\small{Self-Supervised Method:}}}} \\
        
       EpipolarTrans~\cite{he2020epipolar} & 11.2 & 9.0 \\
       LT-Algebraic \cite{iskakov2019learnable}   & 10.3 & 7.8 \\
       LT-Volumetric \cite{iskakov2019learnable}  & 10.6 & 8.0  \\
       LT-Volumetric$^{+}$ \cite{iskakov2019learnable}   & 9.5 & 7.2  \\
       CanonPose$^{+}$ \cite{wandt2021canonpose}   & 21.6 & 10.5  \\
       EpipolarPose$^{+}$ \cite{kocabas2019self}   & 17.2 & 8.3  \\
       \rowcolor{lightblue}
       Ours (Opt-Center)    & 8.8 & \textbf{5.3}  \\
       \rowcolor{lightblue}
       Ours (RANSAC)    & \textbf{8.5} & 5.6  \\
        \bottomrule
      \end{tabular}
      }
      \caption{Multi-view inference results on the HanCo dataset. The notation $^{+}$ indicates that methods require the GT 3D center.}
    \label{tab:multi-view inference}
\end{table}

\begin{table*}[!ht]
\hspace{-0.5cm}
\centering
\makebox[0pt][c]{\parbox{1.05\textwidth}{
\begin{minipage}[h]{0.7\textwidth}
    \centering
    \begin{minipage}[t]{0.9\textwidth}
        \centering
        \setlength{\abovecaptionskip}{0.3cm}
        \scalebox{0.8}{
        
            \begin{tabular}{clccc|cccc}
                \toprule
                \multirow{2}*{ID} &\multirow{2}*{Method} & \multicolumn{3}{c}{NMPJPE~$\downarrow$} &  \multicolumn{3}{c}{PA-MPJPE~$\downarrow$}   
                \\
                \cmidrule(lr){3-5}
                \cmidrule(lr){6-8}
                 & & Single & Interact & Fusion & Single & Interact & Fusion  \\
                 \midrule
    & & & \multicolumn{3}{l}{\emph{\textcolor{lightgray}{\small{ResNet-50 as the backbone:}}}} \\
    1 & Full      & \textbf{11.14}$_{\uparrow  \textcolor[RGB]{20, 150, 20}{0.03}}$ & 8.31$_{\downarrow  \textcolor[RGB]{100, 100, 100}{0.03}}$ & \textbf{7.65}$_{\uparrow  \textcolor[RGB]{20, 150, 20}{0.10}}$ & \textbf{7.05}$_{\uparrow  \textcolor[RGB]{20, 150, 20}{0.17}}$ & \textbf{5.35}$_{\uparrow  \textcolor[RGB]{20, 150, 20}{0.07}}$ & \textbf{5.34}$_{\uparrow  \textcolor[RGB]{20, 150, 20}{0.06}}$ \\
    \midrule
     
    & & &\multicolumn{3}{l}{\emph{\textcolor{lightgray}{\small{ResNet-18 as the backbone:}}}} \\
    2 & Full      & 11.17$_{\qquad}$ & \textbf{8.28}$_{\qquad}$  & 7.75$_{\qquad}$  & 7.22$_{\qquad}$  &  5.42$_{\qquad}$   & 5.40$_{\qquad}$  \\
    \arrayrulecolor{black!30}\midrule
    3 & -- VSF    & 11.21$_{\downarrow  \textcolor[RGB]{100, 100, 100}{0.04}}$ & 8.49$_{\downarrow  \textcolor[RGB]{100, 100, 100}{0.21}}$ & 7.81 $_{\downarrow  \textcolor[RGB]{100, 100, 100}{0.06}}$ & 7.25 $_{\downarrow  \textcolor[RGB]{100, 100, 100}{0.03}}$ & 5.52 $_{\downarrow  \textcolor[RGB]{100, 100, 100}{0.10}}$ & 5.50 $_{\downarrow  \textcolor[RGB]{100, 100, 100}{0.10}}$ \\
    4 & -- CVA    & 11.31$_{\downarrow  \textcolor[RGB]{100, 100, 100}{0.14}}$ & 8.45$_{\downarrow  \textcolor[RGB]{100, 100, 100}{0.17}}$  & 7.81 $_{\downarrow  \textcolor[RGB]{100, 100, 100}{0.06}}$  & 7.29 $_{\downarrow  \textcolor[RGB]{100, 100, 100}{0.07}}$ & 5.48$_{\downarrow  \textcolor[RGB]{100, 100, 100}{0.06}}$ & 5.46$_{\downarrow  \textcolor[RGB]{100, 100, 100}{0.06}}$ \\
    \midrule
    5 & -- $\bm{G}^1$     & 11.31$_{\downarrow  \textcolor[RGB]{100, 100, 100}{0.14}}$ & 8.56$_{\downarrow  \textcolor[RGB]{100, 100, 100}{0.28}}$  & 7.77$_{\downarrow  \textcolor[RGB]{100, 100, 100}{0.03}}$  & 7.31 $_{\downarrow  \textcolor[RGB]{100, 100, 100}{0.09}}$ & 5.52 $_{\downarrow  \textcolor[RGB]{100, 100, 100}{0.10}}$ & 5.49$_{\downarrow  \textcolor[RGB]{100, 100, 100}{0.09}}$ \\
    6 & -- $\bm{G}^2$  & 11.33$_{\downarrow  \textcolor[RGB]{100, 100, 100}{0.16}}$ & 8.38$_{\downarrow  \textcolor[RGB]{100, 100, 100}{0.10}}$  & 7.83$_{\downarrow  \textcolor[RGB]{100, 100, 100}{0.08}}$  & 7.34$_{\downarrow  \textcolor[RGB]{100, 100, 100}{0.08}}$ & 5.45$_{\downarrow  \textcolor[RGB]{100, 100, 100}{0.03}}$ & 5.42$_{\downarrow  \textcolor[RGB]{100, 100, 100}{0.02}}$ \\
    7 & -- $\bm{G}^3$  & 11.30$_{\downarrow  \textcolor[RGB]{100, 100, 100}{0.13}}$  & 8.99$_{\downarrow  \textcolor[RGB]{100, 100, 100}{0.69}}$  & 7.82$_{\downarrow  \textcolor[RGB]{100, 100, 100}{0.07}}$  & 7.30$_{\downarrow  \textcolor[RGB]{100, 100, 100}{0.08}}$ & 5.45$_{\downarrow  \textcolor[RGB]{100, 100, 100}{0.03}}$ & 5.44$_{\downarrow  \textcolor[RGB]{100, 100, 100}{0.04}}$ \\
    
    \midrule

    8 & -- $L_{c_{2D}}$    & 11.25 $_{\downarrow  \textcolor[RGB]{100, 100, 100}{0.08}}$ & 8.43$_{\downarrow  \textcolor[RGB]{100, 100, 100}{0.15}}$ & 7.90$_{\downarrow  \textcolor[RGB]{100, 100, 100}{0.15}}$  & 7.32$_{\downarrow  \textcolor[RGB]{100, 100, 100}{0.10}}$  & 5.58$_{\downarrow  \textcolor[RGB]{100, 100, 100}{0.16}}$ & 5.57 $_{\downarrow  \textcolor[RGB]{100, 100, 100}{0.17}}$\\

    9 & -- $L_{c_{f}}$    & 11.74$_{\downarrow  \textcolor[RGB]{100, 100, 100}{0.57}}$ & 8.98$_{\downarrow  \textcolor[RGB]{100, 100, 100}{0.70}}$  & 8.38$_{\downarrow  \textcolor[RGB]{100, 100, 100}{0.63}}$ & 7.55$_{\downarrow  \textcolor[RGB]{100, 100, 100}{0.33}}$  & 5.84$_{\downarrow  \textcolor[RGB]{100, 100, 100}{0.42}}$  & 5.80$_{\downarrow  \textcolor[RGB]{100, 100, 100}{0.40}}$ \\
    \midrule
    10 & -- DCVI     & 13.52$_{\downarrow  \textcolor[RGB]{100, 100, 100}{2.35}}$ & / & 11.99$_{\downarrow  \textcolor[RGB]{100, 100, 100}{4.24}}$ & 9.59$_{\downarrow  \textcolor[RGB]{100, 100, 100}{2.37}}$ & /  & 9.42$_{\downarrow  \textcolor[RGB]{100, 100, 100}{4.02}}$ \\
    11 & -- $L_{c}$  & 14.04$_{\downarrow  \textcolor[RGB]{100, 100, 100}{2.87}}$  & 17.03$_{\downarrow  \textcolor[RGB]{100, 100, 100}{8.75}}$  & 10.32$_{\downarrow  \textcolor[RGB]{100, 100, 100}{2.57}}$  & 9.04$_{\downarrow  \textcolor[RGB]{100, 100, 100}{1.82}}$  & 10.21$_{\downarrow  \textcolor[RGB]{100, 100, 100}{4.79}}$  &7.92$_{\downarrow  \textcolor[RGB]{100, 100, 100}{2.52}}$   \\ 
    12 & -- $L_{d}$    & 17.05$_{\downarrow  \textcolor[RGB]{100, 100, 100}{5.88}}$ & 8.56$_{\downarrow  \textcolor[RGB]{100, 100, 100}{0.28}}$ & 8.01$_{\downarrow  \textcolor[RGB]{100, 100, 100}{0.26}}$ & 10.13$_{\downarrow  \textcolor[RGB]{100, 100, 100}{2.91}}$ & 5.67$_{\downarrow  \textcolor[RGB]{100, 100, 100}{0.25}}$  & 5.65$_{\downarrow  \textcolor[RGB]{100, 100, 100}{0.25}}$ \\
                             
    \arrayrulecolor{black}\bottomrule
            \end{tabular}}
            \caption{Quantitative ablation studies. We remove each of our components here to show their contribution to our framework. Full denotes our complete model. CVI represents our whole cross-view interaction network. Other notations are consistent with \cref{sec:method}. We report the errors of single-view outputs (Single, $\bm{M}$), cross-view interaction outputs (Interact, $\bm{M}^{*}$), and multi-view fusion results (Fusion, $\tilde{\bm{M}}$). }
            \label{tab:ablation}
    \end{minipage}
\end{minipage}
\hspace{-0.5cm}
\begin{minipage}[h]{0.35\textwidth}
\setlength{\abovecaptionskip}{0cm}
	\centering
	\includegraphics[width=1.05\linewidth]{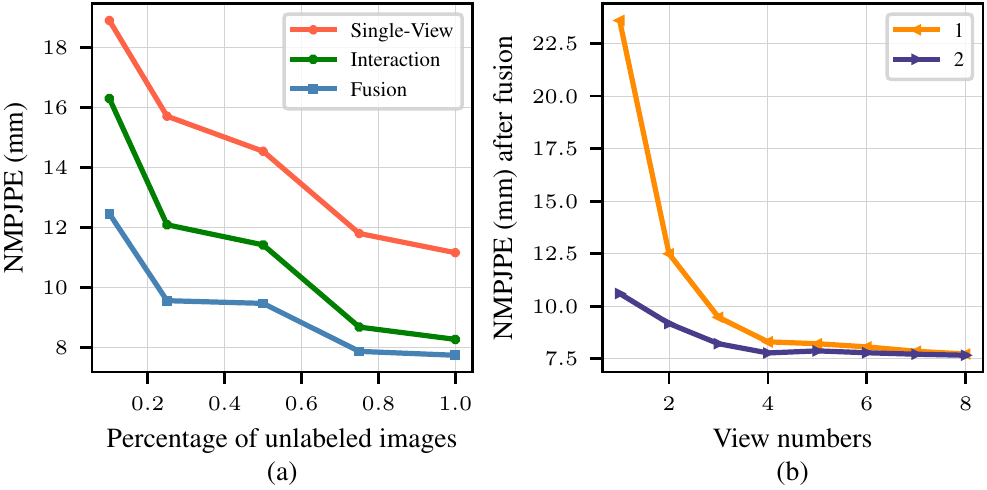}
 	\captionof{figure}{Error of using different (a) \#training data, (b)\;(\textcolor[RGB]{255, 140, 0}{\textbf{line-1}})\#view for training , and (\textcolor[RGB]{102, 51, 153}{\textbf{line-2}})\#view for inference when trained with 8 views.}
	\label{img:percentage data and view}
	\centering
	\includegraphics[width=1.05\linewidth]{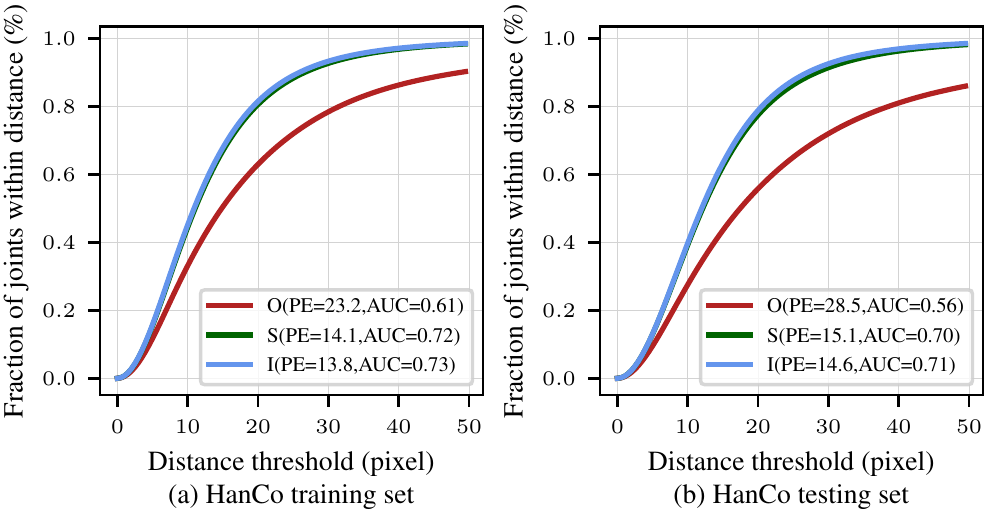}
 	\captionof{figure}{AUC of three 2D joint sets. O, S, I, PE denote OpenPose, single-view, interaction, and average pixel error in resolution $256 \times 256$.}
	\label{img:auc}
\end{minipage}
}}
\vspace{-0.3cm}
\end{table*}

\subsection{Datasets and Metrics}
\noindent{\textbf{FreiHAND}~\cite{zimmermann2019freihand} is a dataset for single-view 3D hand pose estimation, which contains 130,240 training images and 3,960 testing images. All images are captured from the real world with 3D annotations. The training set consists of 32,560 composited images with four types of real-world backgrounds and hands captured against a green screen.
}

\noindent{\textbf{HanCo}~\cite{zimmermann2021contrastive} extends FreiHAND, which consists of 1,517 videos with multiple views
and camera calibration. It has 860,304 frames in total, \ie 107,538 time-step per view. 
Since HanCo does not have an official train/test split, we use the first 1,200 sequences for training and the last 317 sequences for testing in all experiments for fair comparisons.

\noindent
\textbf{Other datasets.} We also provide additional results on other datasets. Assembly101~\cite{sener2022assembly101} is an action recognition dataset that consists of 4,321 videos sequence. H2O~\cite{kwon2021h2o} is a hand-object interaction dataset with 571,645 frames. Please refer to supplementary materials for details.

\noindent \textbf{Metrics.} We report standard metrics for hand pose estimation as follows.
(1) \textbf{MPJPE/MPVPE} (mean per joint/vertex position error) measures the average Euclidean distance in mm between the predicted and ground-truth joints/vertices. JE/VE are the abbreviations for MPJPE/MPVPE.
(2) \textbf{NMPJPE/NMPVPE} (normalized mean per joint/vertex position error, N-JE/VE) computes MPJPE/MPVPE after performing translation and scale alignment.
(3) \textbf{PA-MPJPE/PA-MPVPE}~(PA-JE/VE) is a modification of MPJPE/MPVPE
with Procrustes analysis~\cite{gower1975generalized}. This metric normalizes the absolute scale, center, and rotation. 
(4) \textbf{F-Score}~\cite{chen2021model} is the harmonic mean of recall and precision between two meshes \wrt a specific distance threshold. F@5mm and F@15mm are reported. 
(5) \textbf{AUC} means the area under the curve of the PCK, where the PCK refers to the percentage of correct joints.

\subsection{Implementation Details}
We implement all the networks in PyTorch~\cite{paszke2017automatic}.
We first train our framework without $L_c$ and $L_d$ for 10 epochs. Then, we train the whole framework for another 30 epochs.
Each batch contains images from 8 time-step of 8 cameras.
We use AdamW \cite{loshchilov2017decoupled} optimizer and set the initial learning rate to 3e-4.
We use 256$\times$256 hand images as input.
Please refer to supplementary materials for more details.

\subsection{Comparisons with state-of-the-arts}
In \cref{sec:svi}, we evaluate the performance of our method under the single-view inference setting. As self-supervised hand pose estimation is a relatively new task, there is limited literature available for comparison. To address this, we adapt self-supervised body pose estimation methods~\cite{kocabas2019self,wandt2021canonpose} to hand and compare them with our method on HanCo~\cite{zimmermann2021contrastive}. We then compare with the only existing self-supervised hand pose estimation method, S$^2$Hand~\cite{chen2021model}.
As S$^2$Hand can only be trained on single-view images, we use our single-view network only (denote as Ours-SV) for both training and inference as baselines.
We further conduct extensive evaluations of our full model and baselines to demonstrate the efficacy of multi-view collaborative learning.

In addition, thanks to our cross-view interaction network, our approach is capable of performing multi-view inference by simply averaging individual view results when multi-view test data is available. In \cref{sec:mvi}, we compare our method with state-of-the-art approaches under the multi-view inference setting.

\vspace{-1em}
\subsubsection{Single-View Inference}\label{sec:svi}
\vspace{-0.5em}

\noindent
\textbf{Hanco}. 
We train EpipolarPose and CanonPose using their open-source code. We also train fully-supervised methods~\cite{chen2022mobrecon,kocabas2019self} as a reference for performance.
\cref{tab:single-view inference - hanco} outlines the performance of fully-/self-supervised methods in the literature along with ours. In the case where camera extrinsics are available for training, CanonPose performs the worst because it lifts noisy 2D pseudo labels from OpenPose to 3D ones. When camera extrinsics are not available, all competitors experience a performance decline. 
This is due to the lack of collaborative interaction across multi-view features in previous self-supervised methods.
In contrast, our method outperforms both of them by a large margin. Our cross-view interaction networks can enhance single-view inference, whether camera extrinsics are available during training or not.
More details about the usage of cameras can be found in  \cref{sec:mvc}.
Compared to previous self-supervised methods, our approach significantly improves performance, highlighting the importance of cross-view interaction among different views.
Moreover, our approach can get comparable results to fully-supervised methods.

\noindent
\textbf{FreiHAND}. The comparisons on the evaluation set are shown in \cref{tab:single-view inference - freihand}.
The experiments conducted under self-supervised settings indicate that our baselines, Ours-SV, already achieve performance comparable to S$^2$Hand.
Moreover, directly equipping baselines with other backbones or more training data does not improve too much.
We argue that performance improvements in single-view self-supervised hand pose estimation cannot be achieved by changing the backbone architecture or increasing the amount of training data. 
In contrast, our full model, \ie Ours, substantially further improves the results on the FreiHand dataset, which justify the effectiveness of multi-view collaborative learning.
Moreover, our self-supervised approach achieves competitive performance with recent fully-supervised state-of-the-art methods

\vspace{-1.5em}
\subsubsection{Multi-View Inference}\label{sec:mvi}
\vspace{-0.5em}
\label{exp:multi-view inference}

We show the quantitative results of our multi-view inference performance with other competitors on HanCo in \cref{tab:multi-view inference}. 
A naive solution is to triangulate pseudo labels without training. We show the performance of traditional methods. Such methods can serve as a reference for evaluating the effectiveness of self-supervised methods.
We adapt fully-supervised multi-view 3D pose estimation methods LT\cite{iskakov2019learnable} and EpipolarTrans~\cite{he2020epipolar} to a self-supervised manner.
Under self-supervised settings, EpipolarTrans can only achieve limited performance improvements compared to traditional methods.
LT-Algebraic~\cite{iskakov2019learnable}, which incorporates learnable confidence into the triangulation. LT-Volumetric model \cite{iskakov2019learnable}, which unprojects 2D features into a 3D volume for inference, achieves better results, but the performance is dependent on the accuracy of the hand center.
CanonPose~\cite{wandt2021canonpose} and EpipolarPose~\cite{kocabas2019self} obtain multi-view inference results through simple averaging like ours. 

However, both of these methods are inferior to ours because they lack cross-view interaction.
As our method predicts the root-relative 3D pose, we need to conduct post-processing to obtain the absolute coordinates.
We introduce two different ways to achieve this: 1) using the 2D predictions of different views to triangulate and refine a center and 2) conducting RANSAC triangulation using our 2D predictions.
Both methods have their merits. Opt-center can keep the root-relative results with hand prior, resulting in low PA-MPJPE. RANSAC gets better joint-wise accuracy, which is indicated by low MPJPE.
We also provide qualitative results in the supplementary materials on the Assembly101~\cite{sener2022assembly101} dataset, which has a static camera setup.
Even for challenging head-mounted moving cameras, we achieve convincing 3D pose estimates on the H2O~\cite{kwon2021h2o} dataset.
The experiments show that we have significantly pushed the performance of self-supervised methods to a comparable level with fully supervised methods.

\subsection{Qualitative Result}
\cref{img:2d-comp} presents the visual comparisons of 2 views between 2D joints of OpenPose, ours, and ground-truth on the HanCo dataset. We can observe that our method is more robust for outliers and can generate predictions close to the labels. \cref{img:3d-comp} shows the 3D predictions from two viewpoints of ours, EpipolarPose, and CanonPose on the HanCo dataset. The results indicate that our method can get more accurate results especially when the occlusions are severe.
Please refer to supplementary materials for more results.

\subsection{Ablation Study}
As shown in \cref{tab:ablation}, we conduct comprehensive ablation experiments on the HanCo~\cite{zimmermann2021contrastive} dataset to show the effectiveness of each component. 
Single, Interact and Fusion denotes the evaluation of $\bm{M}$, $\bm{M}^{*}$ and $\tilde{\bm{M}}$ respectively.

\noindent
\textbf{Different backbones.} 
We first show our performance with different backbones. As shown in \#1 and \#2, using a large backbone like Res50, our performance can be further improved. For efficiency, we conduct ablation studies using Res18 as the backbone unless otherwise specified.

\noindent
\textbf{Two branches for cross-view interaction module.} 
As presented in \#3 and \#4, both of the branches can reduce the error.
VSF can explicitly model the view-shared information and add reliable information from every view.
CVA can capture the self-/cross-view joint-level correlations.

\noindent
\textbf{Graph features.}
The results indicate that three kinds of features~(\#5, \#6, \#7) all lead to performance improvement. Especially, local feature (\#7, $\bm{G}^3$) can notably reduce the error after the interaction by providing fine-grained details.

\noindent
\textbf{DCVI.} 
We also conduct experiments to show the importance of DCVI by removing it and posing consistency constraints in single-view outputs like \cite{iqbal2020weakly}. In this way, the performance drops dramatically (\#10), proving the necessity of using DCVI to capture the features of all the views for self-supervised learning.

\noindent
\textbf{Two branches for multi-view consistency loss.} 
Without enforcing cross-view interaction outputs to be consistent, the performance significantly drops (\#9). If we do not explore relatively more reliable 2D predictions to enhance consistency, the performance can also get worse (\#8).

\noindent
\textbf{Consistency losses. ($L_c$)}
The performance is unsatisfactory~(\#11) when employing the cross-view interaction network without any consistency constraints~(\ie discard \#8 and \#9). The interaction network should cooperate with consistency so that the constraints can guide the network to exploit multi-view information to function better.

\noindent
\textbf{Multi-view distillation loss. ($L_d$)} 
Removing the multi-view distillation loss, all the metrics drop by a large margin (\#12), especially in single-view estimation accuracy. This phenomenon proves the effectiveness of collaborative learning between single- and multi-view networks.

\subsection{Model Analysis}

\noindent
\textbf{Different percentage of unlabeled images.} \cref{img:percentage data and view}~(a) shows our method can get consistent performance improvement as the unlabeled training data increases.

\noindent
\textbf{Different view number for training.} 
The \textcolor[RGB]{255, 140, 0}{\textbf{line-1}} in \cref{img:percentage data and view}~(b) shows the performance of our method tested on a certain view when trained with different view numbers. The curve shows that our method can be consistently improved as the number of views increases. We also observe that using multiple views for training can significantly improve performance when the valid views are few.

\noindent
\textbf{Different view number for inference.} 
Our model allows inferring with an arbitrary number of views. However, when the model is trained with a fixed view number, it could get the view number bias, resulting in better performance using the view number close to the training one. To avoid this, we add random masks in our interaction module and finetune the model for a few epochs. After that, results can get better by a small margin (the single-view error is 11.07mm and the fusion error 7.60mm, both in NMPJPE.).
The \textcolor[RGB]{102, 51, 153}{\textbf{line-2}} in \cref{img:percentage data and view}\;(b) shows results on a certain view when trained on 8 views and tested on 1 to 8 views. We can observe consistent improvement with the inference view number increases.

\noindent
\textbf{Different 2D joint sets.} 
\cref{img:auc} presents the accuracy of different 2D joint sets on the HanCo training and testing set. Our 2D predictions are extremely better than OpenPose 2D pseudo label used for training. 

\noindent
\textbf{Iteratively training.}
Our approach can use the previous predictions as pseudo labels for iterative training. We find it helpful till iteration 3 and get saturated afterward. From 1 to 3 iterations, NMPJPE is 7.75, 7.68, and 7.64.

\begin{figure}[t]
	\centering
	\includegraphics[width=0.9\columnwidth]{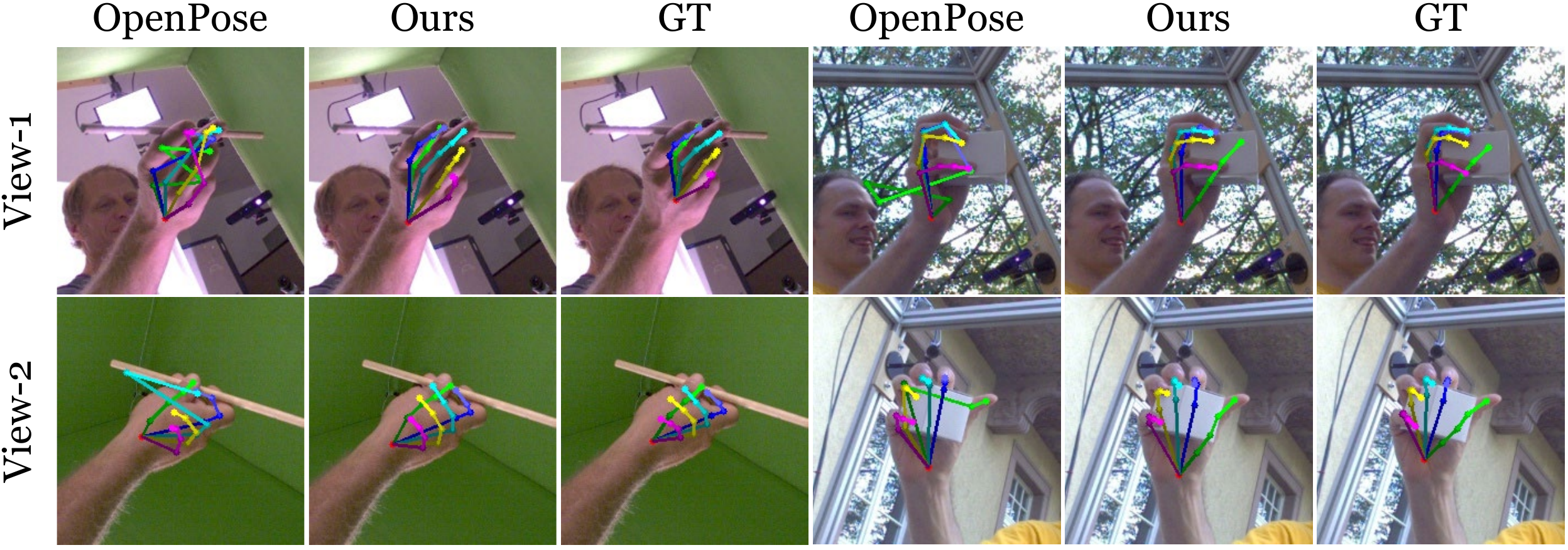} 
	\caption{2D prediction (overlayed in the images) comparisons between OpenPose, ours, and ground-truth on the HanCo dataset.}
	\label{supp:img:2d-comp}
\end{figure}

\begin{figure}[t]
	\centering
	\includegraphics[width=0.9\columnwidth]{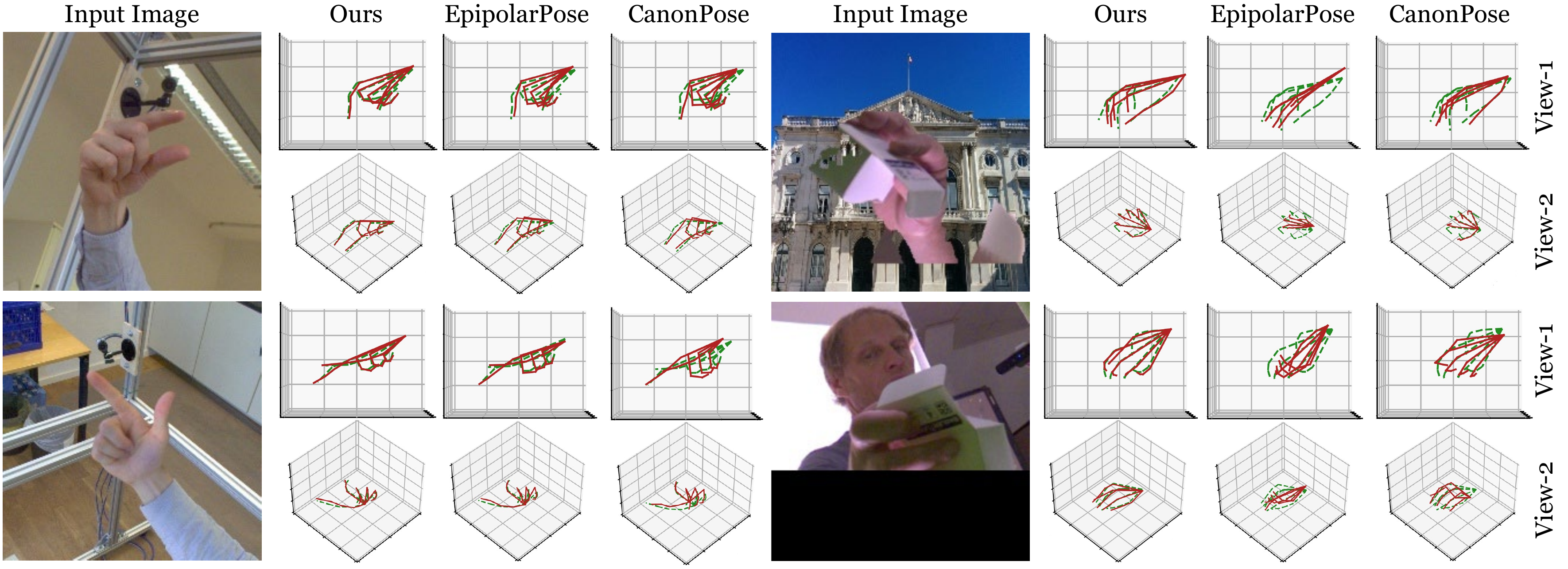} 
	\caption{3D prediction comparisons between our method, EpipolarPose, and CanonPose on the HanCo dataset. Our prediction and ground-truth are shown in solid red and dashed green respectively.
 }
	\label{img:3d-comp}
\end{figure}

\vspace{.4em}
\section{Conclusion and Future Work}
To our best knowledge, we present the first self-supervised framework that aims to learn a single-view 3D hand estimator from unlabeled multi-view data. At the core of our approach, a cross-view interaction network is carefully designed to supervise the single-view output by leveraging the collaboration among multi-views. Specifically, the network captures the interdependencies of features among different views, resulting in improved accuracy of hand pose estimation after cross-view interaction. Additionally, the multi-view results are fused to supervise the single-view output for self-distillation. The effectiveness and versatility of the proposed framework are extensively evaluated through experiments, which demonstrate that our method not only establishes a new benchmark for self-supervised 3D hand pose estimation from single-view input but also offers flexible multi-view inference with state-of-the-art performance.

We focused on hand pose estimation without heavy occlusions in this work.
Extending our work to more challenging scenarios, such as hand-object interaction or relaxing the synchronization constraints in multi-view inputs, would be interesting topics for further study.

\clearpage

\renewcommand{\thesection}{\Alph{section}}  
\renewcommand{\thetable}{\Alph{table}}  
\renewcommand{\thefigure}{\Alph{figure}}

\setcounter{section}{0}
\setcounter{figure}{0}
\setcounter{table}{0}
\twocolumn[{%
\begin{center}
    \textbf{\Large Supplementary Materials}
    \vspace{3em}
    \centering
    \captionsetup{type=figure}
    \includegraphics[width=0.95\textwidth ]{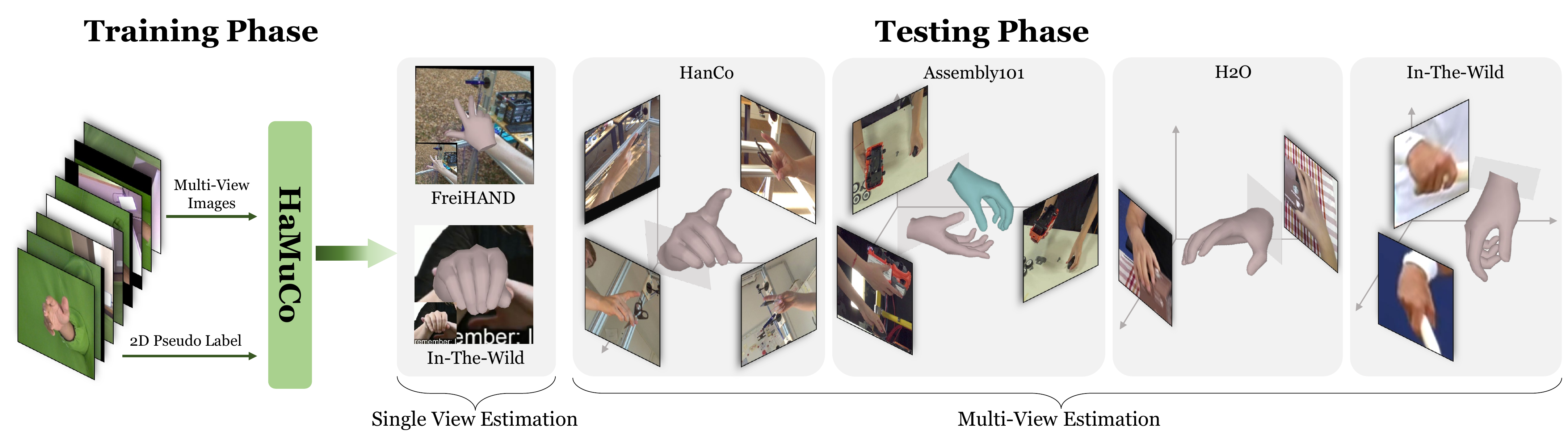}
        \captionof{figure}{Our method takes multi-view images with 2D pseudo labels for training. From the results on public datasets \cite{zimmermann2019freihand,zimmermann2021contrastive,kwon2021h2o,sener2022assembly101} and in-the-wild images, we demonstrate that our method can estimate accurate 3D hand pose with single- or arbitrary multi-view images.}
        \label{img:supp-teaser}
\end{center}%
}]

\maketitle

In the supplemental material, we provide:
\begin{itemize}
\setlength{\itemsep}{0pt}
    \item [\S\ref{suppsec:video}] Video Demo.
    \item [\S\ref{suppsec:details}] Implementation Details.
    \item [\S\ref{suppsec:exp}] More Experiments and Results.
    \item [\S\ref{suppsec:discussions}] Discussions.
\end{itemize}

\section{Video Demo}\label{suppsec:video}
We provide additional sequential qualitative results in the attached video.

\section{Implementation Details}\label{suppsec:details}
\subsection{Single-View Network}

\begin{figure}[!ht]
	\centering
	\includegraphics[width=0.42\textwidth]{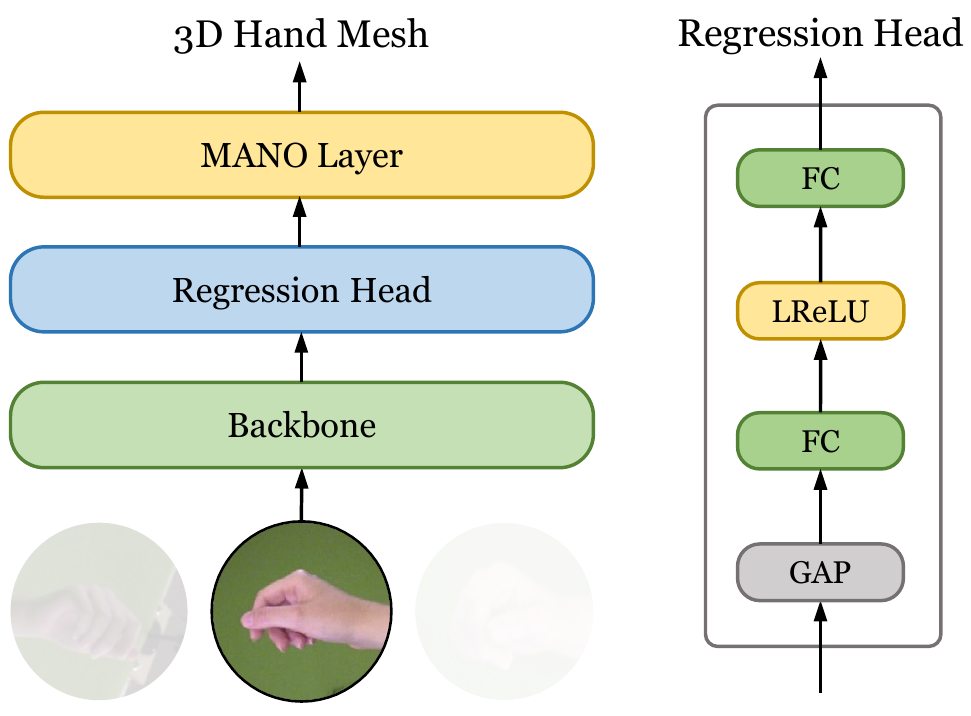}
 	\caption{The details of our single-view estimation network.}
	\label{fig:single-view network} 
\end{figure}

As described in our paper, we only adopt a simple single-view estimation network for our framework. The details of our single-view network are shown in \cref{fig:single-view network}. The network only consists of a backbone (ResNet \cite{he2016deep}) for image feature extraction, a regression head for regressing the MANO \cite{romero2017embodied} parameters, and a MANO layer for parameters decoding to obtain hand mesh. Besides, the regression head is quite simple, only stacking 1 global average pooling (GAP) layer, 2 fully-connected layers, and 1 Leaky-ReLU layer.

\subsection{Multi-View Graph Feature Extraction Module}
Here, we will provide more details about our multi-view graph feature extraction module. The multi-view graph extraction conducts view-shared graph extraction (VSGFE) for each view at first. VSGFE consists of three view-shared modules, a location embedding (LE) module, a spatial-aware initial graph building (SAIGB) module \cite{zheng2021sar}, and a joint feature sampler (JFS). LE uses an MLP to map the predicted 3D joints $\bm{P}_{i} \in \mathbb{R}^{21\times 3}$ and MANO pose parameters (without root joint) $\bm{\theta}^{'}_{i} \in \mathbb{R}^{15\times 3}$ from the single-view estimation network to the joints embeddings $\bm{G}_{i}^{1} \in \mathbb{R}^{21 \times 64}$. SAIGB first uses an MLP to scale the channel number of the high-level feature maps $\bm{H}_{i}^{4} \in \mathbb{R}^{2048 \times 8 \times 8}$ to a dimension $21\times 8$. Then, it reshapes the features to obtain $\bm{G}_{i}^{2} \in \mathbb{R}^{21 \times 512}$. Motivated by ~\cite{wang2018pixel2mesh,wen2019pixel2mesh++}, we design a joint feature sampler (JFS) to sample the joint-aligned features from the middle-level feature maps. The details of our JFS are shown in \cref{fig:joint-feature sampler}.
Given the 3D coordinates of hand joints, we calculate its 2D projections on the feature map using weak perspective projection, then gather the features from nearby pixels via bilinear interpolation.
In particular, we sample the joint-aligned features from three levels of the feature maps $\{  \bm{H}_{i}^{j}\}_{j=1}^{3}$ to obtain $\bm{G}_{i}^{3} \in \mathbb{R}^{21 \times 1792}$. After concatenation and stack, we obtain multi-view graph feature $\bm{G} \in \mathbb{R}^{21 \times 2368}$.

\begin{figure}[!ht]
	\centering
	\includegraphics[width=0.85\columnwidth]{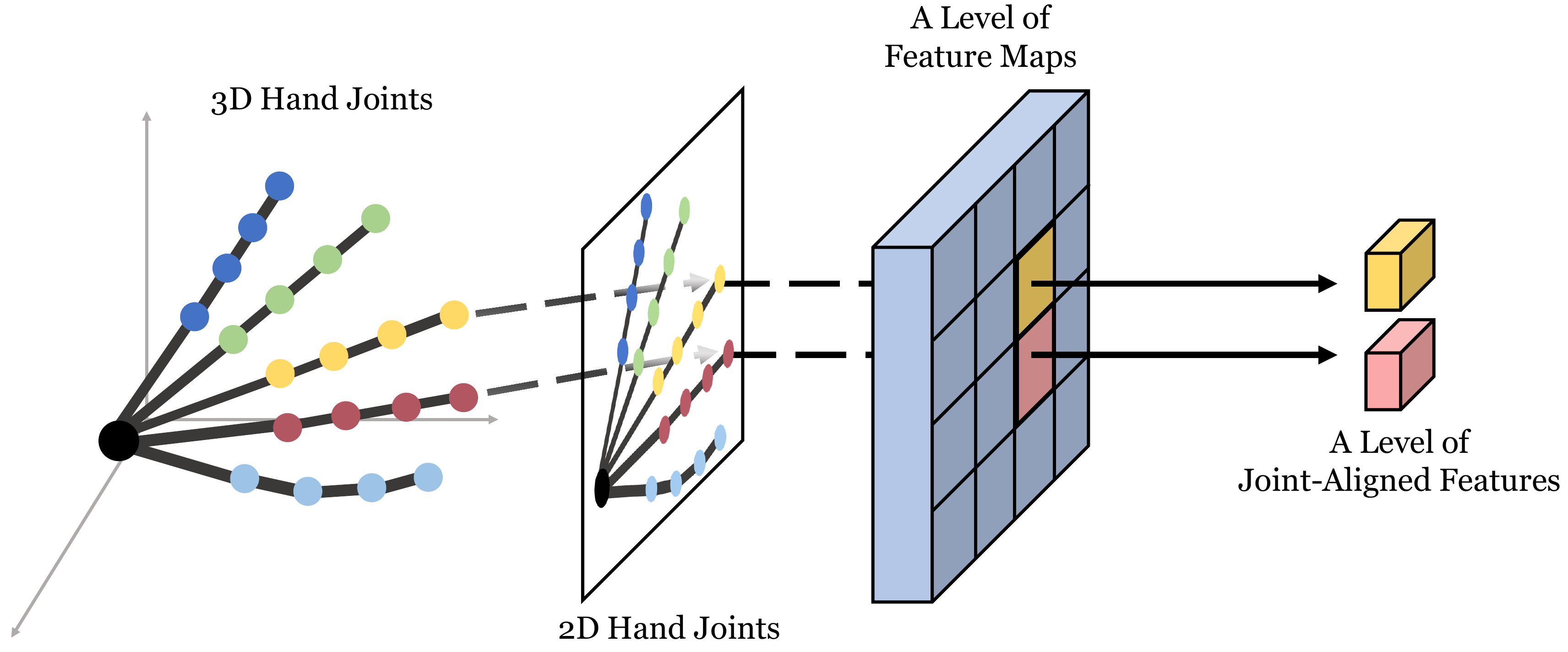}
 	\caption{Illustration of our joint feature sampler (JFS) sampling a level of the joint-aligned features for 2 joints.}
	\label{fig:joint-feature sampler} 
\end{figure}


\begin{table}[!t]
  \centering
  \resizebox{0.97\columnwidth}{!}{
  \begin{tabular}{rrrrl}
    \toprule
    \#Out & \#In & Shape & Operation & Notation \\
    \midrule
    \multicolumn{5}{l}{\emph{\textcolor{lightgray}{\small{Backbone:}}}} \\
    1 & /  & (8, 3, 256, 256) & Input & $\bm{I}$ \\
    2 & 1  & (8, 64, 64, 64) & ResLayer \\
    3 & 2  & (8, 256, 64, 64) & ResBlock1 & $\bm{H^{1}}$ \\
    4 & 3  & (8, 512, 32, 32) & ResBlock2 & $\bm{H^{2}}$ \\
    5 & 4  & (8, 1024, 16, 16) & ResBlock3 & $\bm{H^{3}}$ \\
    6 & 5  & (8, 2048,  8, 8) & ResBlock4 & $\bm{H^{4}}$ \\
    \midrule
    \multicolumn{5}{l}{\emph{\textcolor{lightgray}{\small{Single-View Decoder:}}}} \\
    7 & 6  & (8, 2048) & GAP  \\
    8 & 7  & (8, 48)           & MLP & $\bm{\theta}$ \\
    9 & 7  & (8, 10)           & MLP & $\bm{\beta}$\\
    10 & 7  & (8, 3)           & MLP & $s,\bm{t}$\\\
    11 & 8,9  & (8, 778, 3)     & MANO & $\bm{M}$ \\
    12 & 11  & (8, 21, 3)      & Regressor & $\bm{P}$ \\
    \midrule
    \multicolumn{5}{l}{\emph{\textcolor{lightgray}{\small{Multi-View Graph Feature Extraction:}}}} \\
    13 & 8,12  & (8, 21, 64)  & LE & $\bm{G}^{1}$\\
    14 & 6  & (8, 21, 512) & SAIGB & $\bm{G}^{2}$\\
    15 & 3,4,5  & (8, 21, 1792) & JFS & $\bm{G}^{3}$ \\
    16 & 13,14,15  & (8, 21, 2368) & Concat &  \\
    17 & 16  & (168, 2368) & Reshape & $\bm{G}$ \\
    \midrule
    \multicolumn{5}{l}{\emph{\textcolor{lightgray}{\small{Dual-Branch Cross-View Interaction:}}}} \\
    18 & 17  & (168, 2368) & CVA-1 &  \\
    19 & 17  & (168, 2368) & VSF-1 &  \\
    20 & 17,18,19  & (168, 2368) & Add &  \\
    21 & 20  & (168, 2368) & CVA-2 & $F_{t}(\bm{G})$ \\
    22 & 20  & (168, 2368) & VSF-2 & $\bm{C}^{'}$ \\
    23 & 20,21,22  & (168, 2368) & Add & $\bm{G}^{*}$  \\
    \midrule
    \multicolumn{5}{l}{\emph{\textcolor{lightgray}{\small{Parameters Regression:}}}} \\
    24 & 23  & (168, 32) & MLP & \\
    25 & 24  & (8, 672) & Reshape &  \\
    26 & 25  & (8, 48) & MLP & $\bm{\theta}^{*}$ \\
    27 & 25  & (8, 3) & MLP & $s^{*},\bm{t}^{*}$ \\
    28 & 9,26  & (8, 778, 3) & MANO & $\bm{M}^{*}$ \\
    29 & 28  & (8, 21, 3)    & Regressor & $\bm{P}^{*}$ \\
    
    \bottomrule
  \end{tabular}
  }
  \caption{The architecture of our whole network. We show the output shapes after every operation when adopting ResNet-50 as the backbone and taking 8 views of images of resolution $256\times256$ as the input. \#Out and \#In denotes the output and input index of this operation. In the last column, we specify those outputs that have notations in our paper.
  }
  \label{tab:architecture}
\end{table}

\subsection{Architecture Details}
\cref{tab:architecture} shows the details of our complete architecture. Unless otherwise specified, MLP denotes using 
2 fully-connected layers and 1 Leaky-ReLU layer (same as the regression head in \cref{fig:single-view network} without GAP). We use 2 layers of CVA and VSF in the dual-branch cross-view interaction module (e.g. CVA-1 denotes the first CVA branch).

\subsection{Loss Weights}
To balance multiple loss functions, we introduce $\alpha$ and $\gamma$ in our loss function. For all of our experiments, we set $\alpha=0.01$ and $\gamma=100$. It is worth mentioning that adjusting $\alpha$ to a correct scale is important for self-supervised learning because $\alpha$ balances the strength of hand-prior information provided by the MANO and the trustworthiness of pseudo labels. When the pseudo labels are reliable, we can reduce $\alpha$ to trust the pseudo labels more. Otherwise, we should enlarge $\alpha$ to use MANO to regularize irrational poses.

\subsection{Hand Center Coordinate System}
As shown in \cref{img:supp-teaser}, our method can be used for multi-view inference with or without camera extrinsics.
If the camera extrinsics are known (HanCo \cite{zimmermann2021contrastive} and Assembly101 \cite{sener2022assembly101}), the coordinate system of the hand center is the world coordinate system. 
If the extrinsics are not available (H2O \cite{kwon2021h2o} and in-the-wild), we choose one view as the reference view, and the center is located in this reference view coordinate system.

\section{Experiments and Results}\label{suppsec:exp}

\begin{table}[b]
\centering
\begin{tabular}{c|cccc}
    \hline
    Scheme             & Stage & Intrinsic & Extrinsic & GT Center \\ \hline
    \multirow{2}{*}{1} & Train & $\times$     &  $\times$/$\checkmark$   & $\times$         \\
                       & Test  & $\times$     & $\times$     & $\times$        \\ \hline
    \multirow{2}{*}{2} & Train & $\times$     & $\checkmark$     & $\times$         \\
                       & Test  & $\checkmark$     & $\checkmark$     & $\times$         \\ \hline
    \end{tabular}
    
    \caption{Different assumptions for HaMuCo.}
    \label{tab:assumption}
\end{table}

\subsection{Different Settings}
We show the different assumptions of our experiments in \cref{tab:assumption}.
There are generally two settings, and in both settings, we do not require GT centers. 
For single-view inference, which corresponds to Tab.1 and Tab.2 in the main text. 
Extrinsics are optionally used during the training phase, and all experiments that utilize camera extrinsics are marked with {\footnotesize \faCamera}.
The multi-view inference is an additional benefit of our method, corresponding to Tab.3. 
Only in the test phase, do we require both intrinsic and extrinsic to obtain the 3D pose of absolute scale.

\subsection{Datasets}

\noindent
\textbf{Assembly101} \cite{sener2022assembly101} is an action recognition dataset that consists of 4,321 videos recording different persons manipulating toys. It is recorded by 8 simultaneous static cameras and 4 egocentric cameras. We only use 8 sequences of 8 static cameras for training and present the qualitative results on an additional sequence.

\noindent
\textbf{H2O} \cite{kwon2021h2o} provides synchronized multi-view RGB-D images with two hands manipulating objects.  The data captured by 4 static cameras and 1 egocentric camera consists of 344,645 frames for training,
73,380 frames for validation and 153,620 frames for testing. We only evaluate our cross-dataset performance on this dataset using one sequence with 1 egocentric camera and 2 static cameras.

\subsection{Pseudo Labelling}
We obtain the 2D joints pseudo labels at an offline stage through an implementation\footnote{\url{https://github.com/Hzzone/pytorch-openpose}} of OpenPose~\cite{simon2017hand,cao2017realtime}. For HanCo~\cite{zimmermann2021contrastive}, we directly input the images with the original size due to the images having been cropped already. For Assembly101~\cite{sener2022assembly101}, we use a hand detector to locate and crop the hands. Then, we input the cropped images to obtain the pseudo labels. 

\subsection{Model Analysis}

\noindent
\textbf{Different view number for training and inference.} Here, we explain the camera settings of the experiments evaluating the performance of our models using different view numbers for training and inference~(Fig.~3 in the main submission). Specifically, all the camera settings follow two rules. First, we only test the performance on a specific view for fair comparisons, considering only one specific view is available for all the experimental settings. Second, we choose camera combinations that cover a wider field of vision so that more information can be provided when the camera number has been determined.

\noindent
\textbf{Multi-view weakly-supervised learning.}
Our method can also be applied to weakly-supervised learning.
Therefore, we conduct an experiment to show the performance of our model using weak 2D supervision.
Considering the 2D labels from different views of the HanCo dataset are projected by the same 3D label, using all the 2D labels as weak supervisions may introduce implicit 3D supervision. Therefore, we only utilize the 2D labels from a specific view for weakly-supervised learning. During the training, we set the confidence of the labels to 1. As shown in \cref{tab:1-view 2D-GT}, when incorporating the label of a view, the performance can be improved. The performance improvement of single-view and interaction without alignments is not significant compared to others. The reason may be two folds. First, it is difficult to obtain a correct rotation from single-view inference. Second, multi-view inference without extrinsics is not able to well correct the global rotation error from every single view. In summary, our method can benefit from available 2D labels, especially when using multi-view images for inference.

\begin{table}[t]
  \centering
  \resizebox{0.97\columnwidth}{!}{
  \begin{tabular}{llllll}
    \toprule
    \multicolumn{3}{c}{NMPJPE~$\downarrow$} & \multicolumn{3}{c}{PA-MPJPE~$\downarrow$} \\
    \cmidrule(lr){1-3}
    \cmidrule(lr){4-6}
    Single & Interact & Fusion & Single & Interact & Fusion \\
    \midrule
    \multicolumn{5}{l}{\emph{\textcolor{lightgray}{\small{Self-supervised learning:}}}}\\
    11.17$_{\qquad}$ & 8.28$_{\qquad}$  & 7.75$_{\qquad}$  & 7.22$_{\qquad}$  &  5.42$_{\qquad}$   & 5.40$_{\qquad}$  \\
    \midrule
    \multicolumn{6}{l}{\emph{\textcolor{lightgray}{\small{Weakly-supervised learning (\textcolor{blue}{one view} of the 2D ground-truth is available):}}}} \\
    11.06$_{\uparrow  \textcolor[RGB]{20, 150, 20}{0.11}}$ & 7.84$_{\uparrow  \textcolor[RGB]{20, 150, 20}{0.44}}$  & 6.84$_{\uparrow  \textcolor[RGB]{20, 150, 20}{0.91}}$  & 6.87$_{\uparrow  \textcolor[RGB]{20, 150, 20}{0.35}}$  &  4.49$_{\uparrow  \textcolor[RGB]{20, 150, 20}{0.93}}$   & 4.44$_{\uparrow  \textcolor[RGB]{20, 150, 20}{0.96}}$  \\
    
    \bottomrule
  \end{tabular}
  }
  \caption{Performance comparisons of our method under self- and weak- supervised settings.}
  \label{tab:1-view 2D-GT}
\end{table}

\begin{table}
\centering
 \resizebox{0.97\columnwidth}{!}{
 \begin{tabular}{cccccccc}
            \toprule
            Method                             & Data  & Backbone          & PA-JE$\downarrow$ & PA-VE$\downarrow$ & F@5$\uparrow$ & F@15$\uparrow$ \\
            \midrule
            \multicolumn{3}{l}{\emph{\textcolor{lightgray}{\small{{Fully-Supervised Method:}}}}}
            \\
            YoutubeHand \cite{kulon2020weakly} & FreiHAND &  Res50  & 8.4 & 8.6  & 0.61  & 0.97 \\
            I2L-MeshNet \cite{moon2020i2l} & FreiHAND &  Res50$^{\dagger}$  & 7.4  & 7.6  & 0.68  & 0.97 \\
            METRO \cite{lin2021end} & FreiHAND &  HRNet  & 6.7  & 6.8  & 0.72  & 0.98 \\
            Tang \etal \cite{tang2021towards} & FreiHAND &  Res50  & 6.7  & 6.7  & 0.72  & 0.98 \\
            I2UV-HandNet \cite{chen2021i2uv}& FreiHAND& Res50& 6.7 & 6.9 & 0.71 & 0.98 \\
            MobRecon \cite{chen2022mobrecon}& FreiHAND& Res50$^{\dagger}$ & 6.1 & 6.2 & 0.76 & 0.98 \\
            \rowcolor{lightgray2}
            Ours-SV    & Frei. & Res50 & 7.5               &  7.5        & 0.68   & 0.97      \\
            \midrule
            \multicolumn{3}{l}{\emph{\textcolor{lightgray}{\small{{Weakly-Supervised Method:}}}}}
            \\
            S$^{2}$HAND\cite{chen2021model}    & Frei. & EffiNet-b0        & /                 & /                 & 0.42  & 0.89        \\
            \rowcolor{lightgray2}
            Ours-SV                     & Frei. & EffiNet-b0             & 8.5               & 8.6           & 0.61     &0.97     \\
            \rowcolor{lightgray2}
            Ours-SV                     & Frei. & Res50             & 9.8               & 9.9               & 0.55    & 0.95     \\
            \midrule
            \multicolumn{3}{l}{\emph{\textcolor{lightgray}{\small{Self-Supervised Method:}}}}
            \\
            S$^{2}$HAND\cite{chen2021model}    & Frei. & EffiNet-b0        & 11.8              & 11.9              & 0.48  & 0.92         \\
            \rowcolor{lightgray2}
            Ours-SV                     & Frei. & EffiNet-b0             & 11.6              & 11.7          & 0.49   & 0.93      \\
            \rowcolor{lightblue}
            Ours                               & HanCo & EffiNet-b0             & 6.3              & 6.8              & 0.71   & \textbf{0.99}       \\
            \rowcolor{lightblue}
            Ours                               & HanCo & Res50             & \textbf{6.2}               & \textbf{6.7}               & \textbf{0.72}   & \textbf{0.99}       \\
            \bottomrule
            \end{tabular}
    }
    \caption{Quantitative results on the FreiHAND evaluation set. The notation $^{\dagger}$ denotes using a stacked backbone structure. "Our-SV" refers to training only with our single-view network. 
    }\label{tab:supp_frei}
\end{table}

\subsection{Results for Human Pose Estimation}
Our method can also be extended to self-supervised human pose estimation. 
Therefore, we conduct experiments on the Human3.6M dataset~\cite{ionescu2013human3} to compare with EpipolarPose \cite{kocabas2019self} and CanonPose~\cite{wandt2021canonpose}.
We train our model following the training setting of CanonPose~\cite{wandt2021canonpose}.
When using camera extrinsics for multi-view self-supervised learning, the NMPJPE (mm$\downarrow$) for EpipolarPose, CanonPose, and ours are 76.6, 74.3, and 71.1, respectively.

\subsection{Additional Quantitative Results}
\noindent\textbf{FreiHand.}
\cref{tab:supp_frei} shows more quantitative comparisons between our approach and recent fully-supervised methods. The experimental results demonstrate that our self-supervised method achieves comparable performance to fully supervised methods~\cite{kulon2020weakly,moon2020i2l,lin2021end,tang2021towards,chen2021i2uv,chen2022mobrecon}.
We also compared our method with S$^2$Hand~\cite{chen2021model}, a hand pose estimation method in the weakly supervised setting, which uses annotated 2D labels instead of pseudo labels to estimate 3D results. The experimental results demonstrate that our method is still effective under weak supervision.

\subsection{Additional Qualitative Results}
As illustrated in \cref{img:supp-teaser}, our model is capable of performing inference on multiple datasets~\cite{sener2022assembly101,kwon2021h2o,zimmermann2019freihand,zimmermann2021contrastive}.

\cref{supp:img:2d-comp} shows the 2D visual comparisons between OpenPose, our single-view inference results, and the ground-truth. The results demonstrate that OpenPose can obtain plausible results for those visible joints, which is essential for self-supervised learning. However, the major problem with OpenPose is that it is not robust for invisible joints. When some joints are invisible, it can predict some particularly incorrect results and tend to predict the visible joints as the invisible ones. In contrast, our model-based method with hand prior information obtains a more robust performance towards different kinds of occlusions when the multi-view self-supervised learning provides enough accurate results for supervision. 

\cref{supp:img:3d-comp} provides more visual comparisons between our method, EpipolarPose \cite{kocabas2019self}, and CanonPose \cite{wandt2021canonpose}. All these 3D predictions are obtained with the single-view inference of the models trained by multi-view self-supervised learning. Besides, for better visualization, the predictions in the images are results after alignment with the ground-truth. From the predictions from 2 viewpoints, we can see that our method can obtain more accurate 3D joints with different gestures, backgrounds, viewpoints, occlusions, and objects in hands.

\cref{img:assembly} displays the visualization of our method on the testing sequence of the Assembly101 dataset. We only train a right-hand model, and the left-hand predictions are obtained using the flipped left-hand cropped images for inference. The results demonstrate that our method can be applied to more complicated situations where the available number of hands is unknown at each time step and the occlusions are severe.

\cref{img:learnable-triangulation} compares our multi-view inference performance with Learnable Triangulation \cite{iskakov2019learnable} (algebraic version). All the models are trained with self-supervised learning. The predictions are aligned with the ground-truth for better visualization. The results indicate that our method can generate more plausible results with multi-view inference when the camera parameters are available. 

\cref{img:h2o} illustrates our cross-dataset predictions on the testing sequence of the H2O dataset. We make use of our model trained on the HanCo dataset to estimate the hand poses with images from multiple uncalibrated cameras. The results demonstrate that our method can generalize to other multi-view settings with unknown camera parameters. 

\cref{img:freihand} visualizes the 2D prediction comparisons between S$^{2}$HAND\cite{chen2021model}, our method, and the ground-truth on the evaluation set of the FreiHAND dataset \cite{zimmermann2019freihand}. The results of S$^{2}$HAND are obtained by their open-source code\footnote{\url{https://github.com/TerenceCYJ/S2HAND}} with the provided pretrained weights. As shown in the images, our model using multi-view self-supervised learning on the HanCo dataset can obtain plausible single-view predictions on the FreiHAND dataset. 

\cref{img:failure cases} presents our failure cases on the HanCo dataset. Most of our fails are predictions from samples with challenging viewpoints and severe occlusions. Moreover, the failing predictions mainly fall into two patterns. One is incorrect hand scales and centers, and the other is wrong hand poses. Since the cross-view interaction does not explicitly use the camera extrinsics, it is difficult for it to fix those predictions with incorrect scale and center. However, from those results, we can see that it can solve the incorrect hand poses to some extent.

\section{Discussions}\label{suppsec:discussions}
\subsection{Difference between Qiu \etal~\cite{qiu2019cross} and Ours}
Our cross-view interaction network differs from Qiu \etal~\cite{qiu2019cross} in various aspects.
(1) Regarding motivation, 
our cross-view interaction is designed to generate more reliable results for self-supervision of our single-view network while \cite{qiu2019cross} aims at fusing different views' heatmaps for multi-view inference.
(2) In terms of representation, 
our cross-view interaction utilizes compact and effective joint-level features for dual-branch interaction, while \cite{qiu2019cross} fuses pixel-level features along the epipolar line, which can be computationally expensive.
(3) In terms of usage,
our cross-view interaction does not require camera extrinsics since we fuse information in semantic joint space while \cite{qiu2019cross} relies on extrinsics for finding the epipolar line to do pixel feature fusion.

\begin{figure*}[!t]
	\centering
	\includegraphics[width=0.9\textwidth]{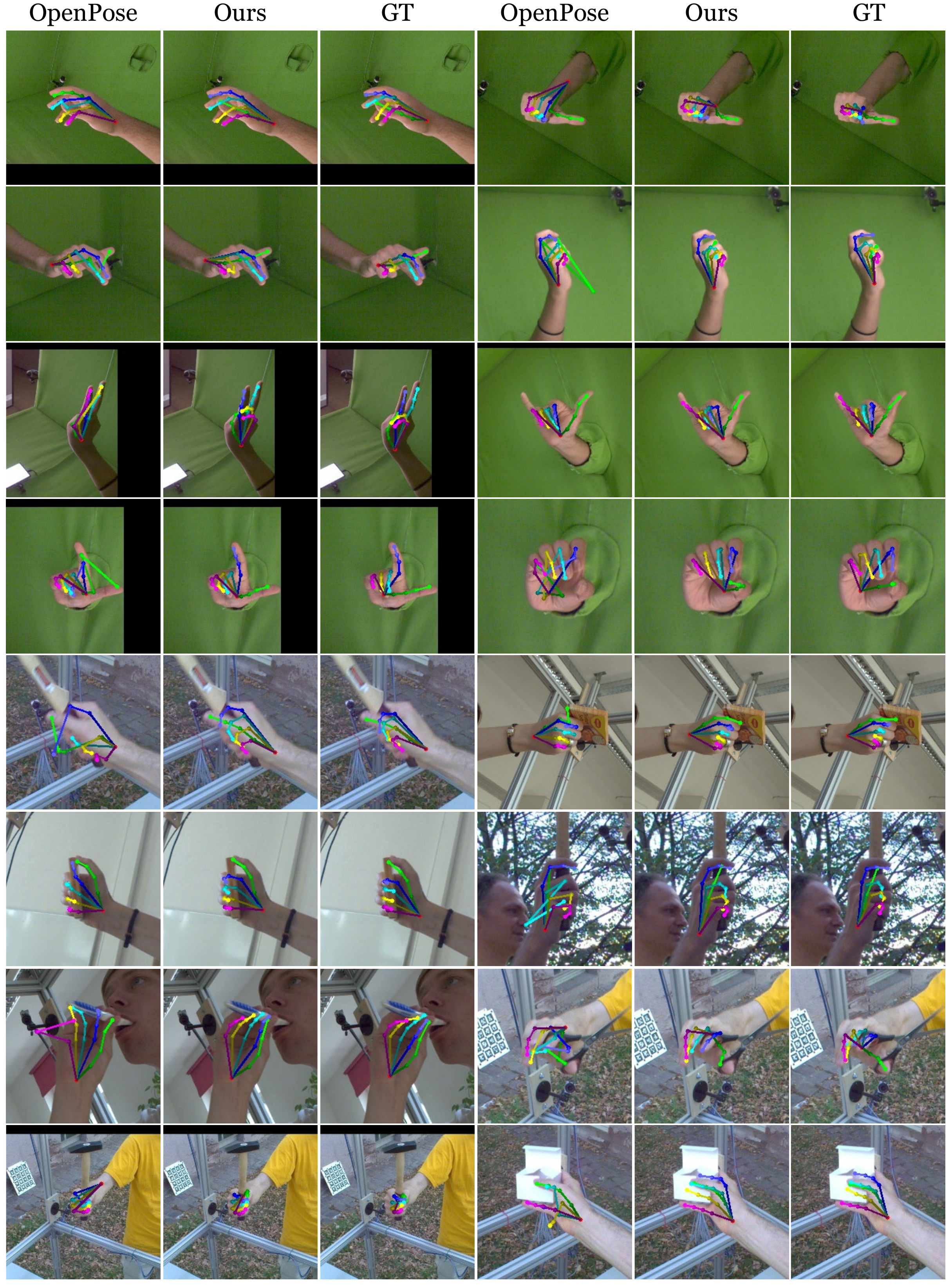} 
	\caption{2D prediction (overlayed in the images) comparisons between OpenPose, ours, and the ground-truth on the HanCo dataset.}
	\label{img:2d-comp}
\end{figure*}

\begin{figure*}[!t]
	\centering
	\includegraphics[width=0.90\textwidth]{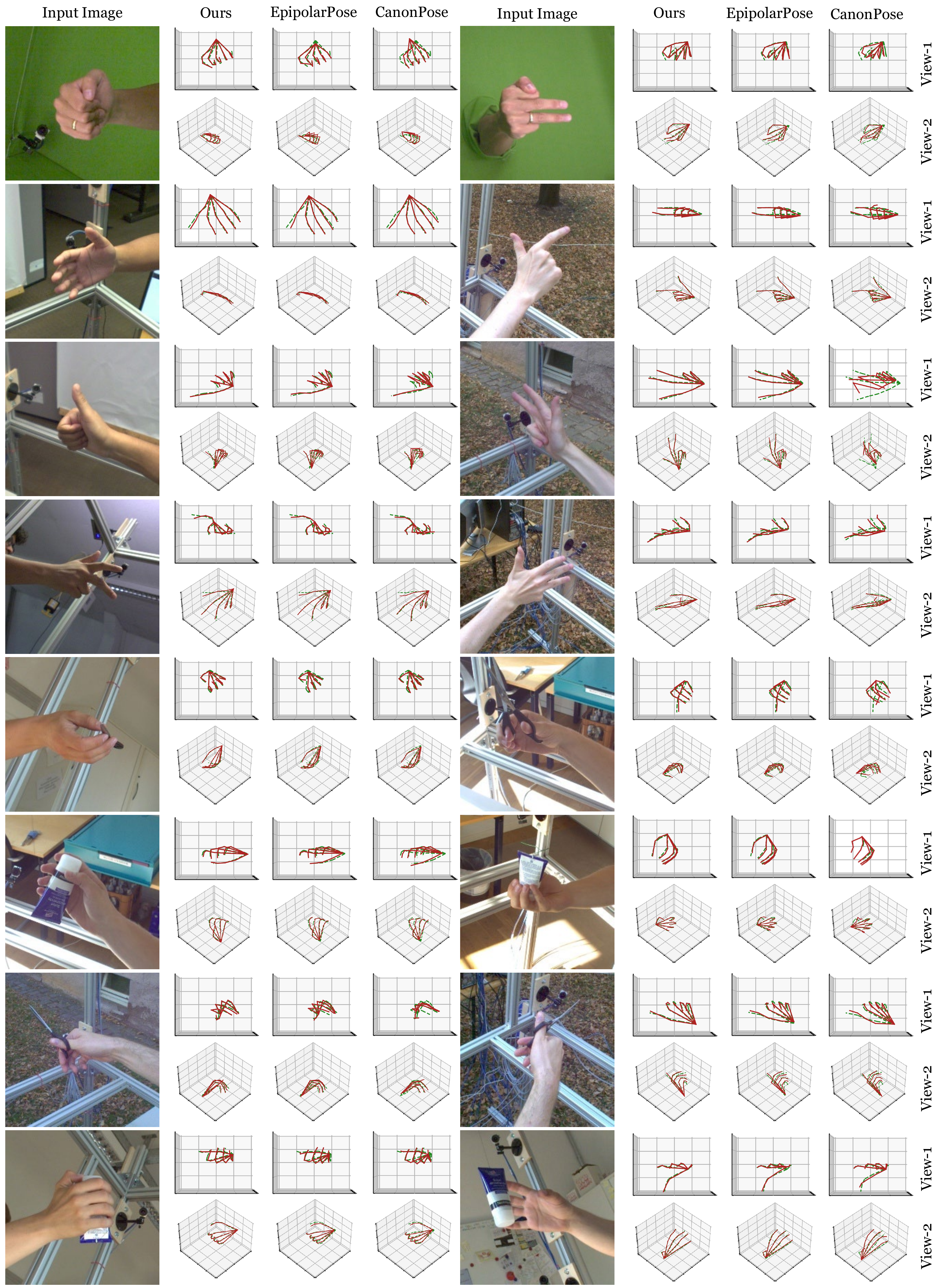} 
	\caption{3D prediction comparisons between our method, EpipolarPose, and CanonPose on the HanCo dataset. Our prediction and the ground-truth are shown in solid red and dashed green respectively.}
	\label{supp:img:3d-comp}
\end{figure*}

\begin{figure*}[!t]
	\centering
	\includegraphics[width=0.75\textwidth]{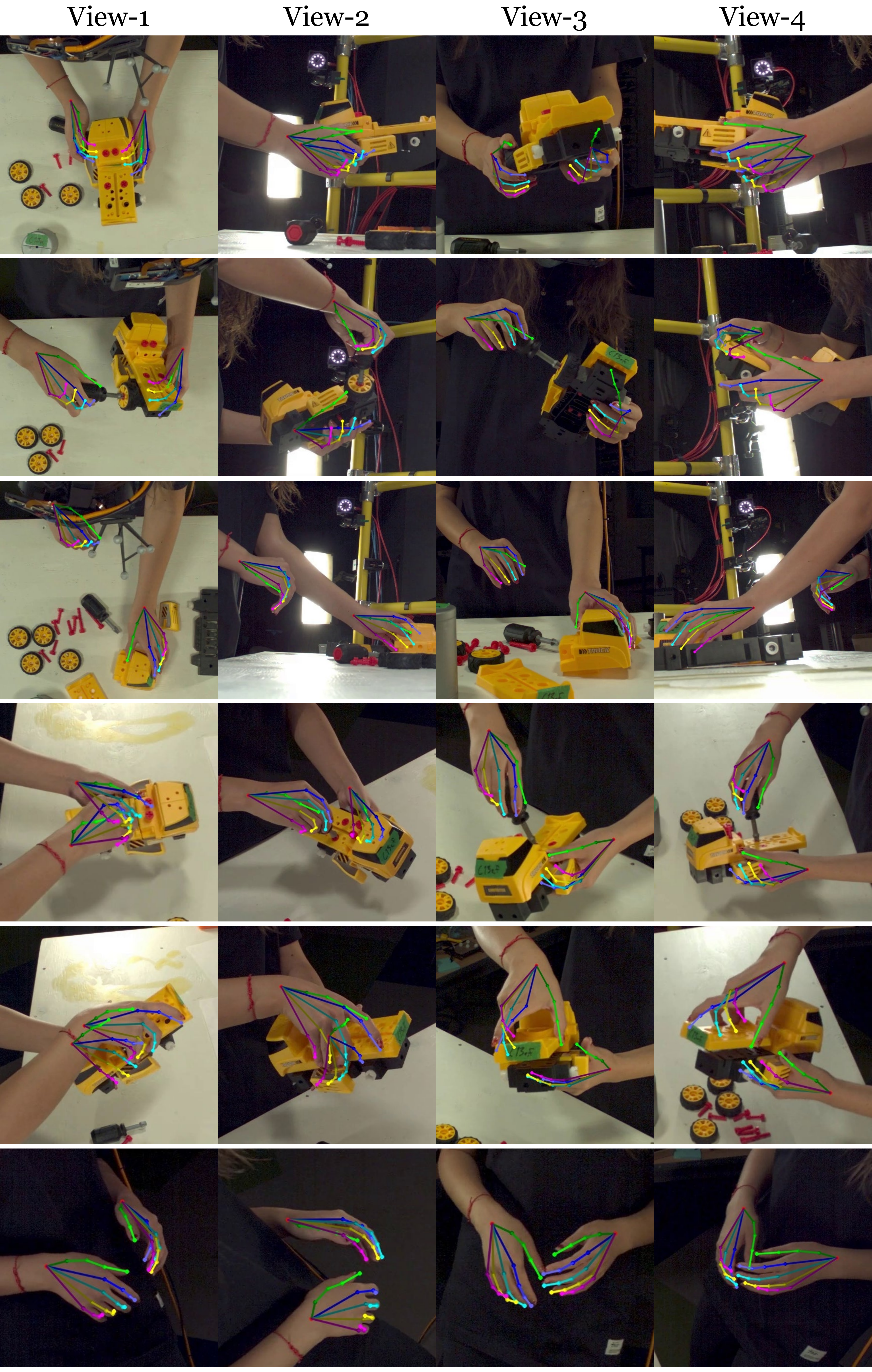} 
	\caption{2D prediction (overlayed in the images) of our method in the testing sequence of the Assembly101 dataset. All the 2D image coordinates are obtained by projecting the same 3D world coordinates into different views. We utilize 8 views in total for inference. Each row shows 4 views of the projected 2D joints. The top 3 rows display the images on 4 views out of all the views, while the bottom 3 rows present the results of another 4 views.}
	\label{img:assembly}
\end{figure*}

\begin{figure*}[!t]
	\centering
	\includegraphics[width=0.95\textwidth]{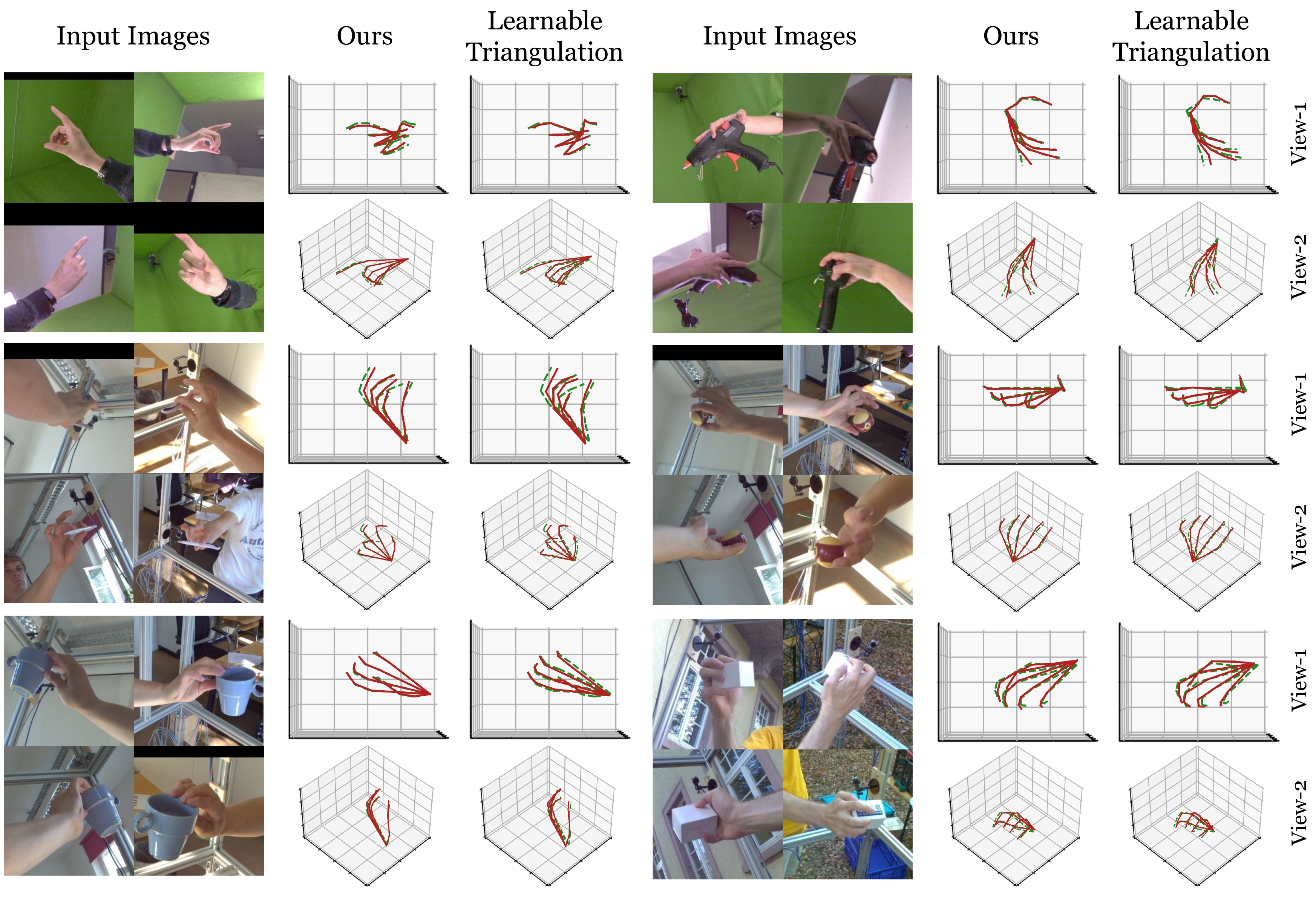} 
	\caption{3D prediction comparisons between our method and Learnable Triangulation on the HanCo dataset. Our prediction and the ground-truth are shown in solid red and dashed green respectively. We use 8 views for inference and only show 4 images here.}
	\label{img:learnable-triangulation}
\end{figure*}

\begin{figure*}[!t]
	\centering
	\includegraphics[width=0.95\textwidth]{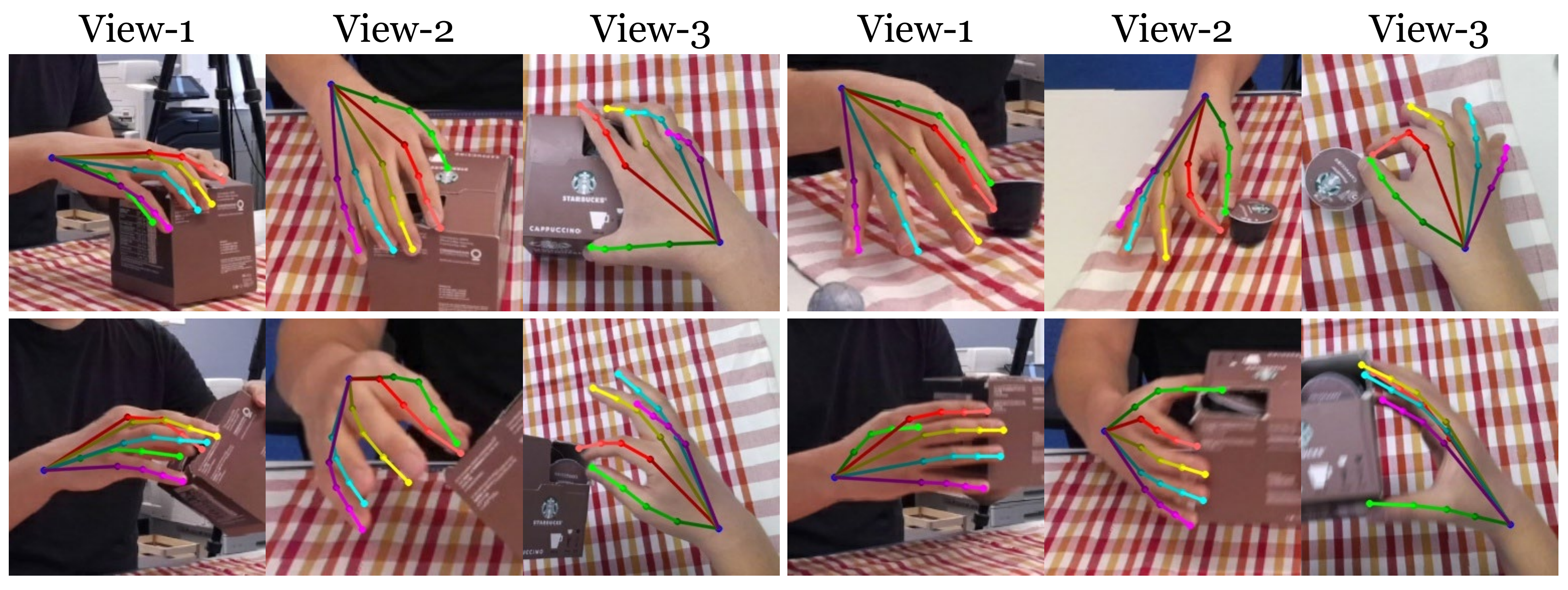} 
	\caption{2D prediction (overlayed in the images) of our method in the testing sequence of the H2O dataset. The results are obtained by the model trained on the HanCo dataset. We use 3 views for inference without camera extrinsics.}
	\label{img:h2o}
\end{figure*}

\begin{figure*}[!t]
	\centering
	\includegraphics[width=0.92\textwidth]{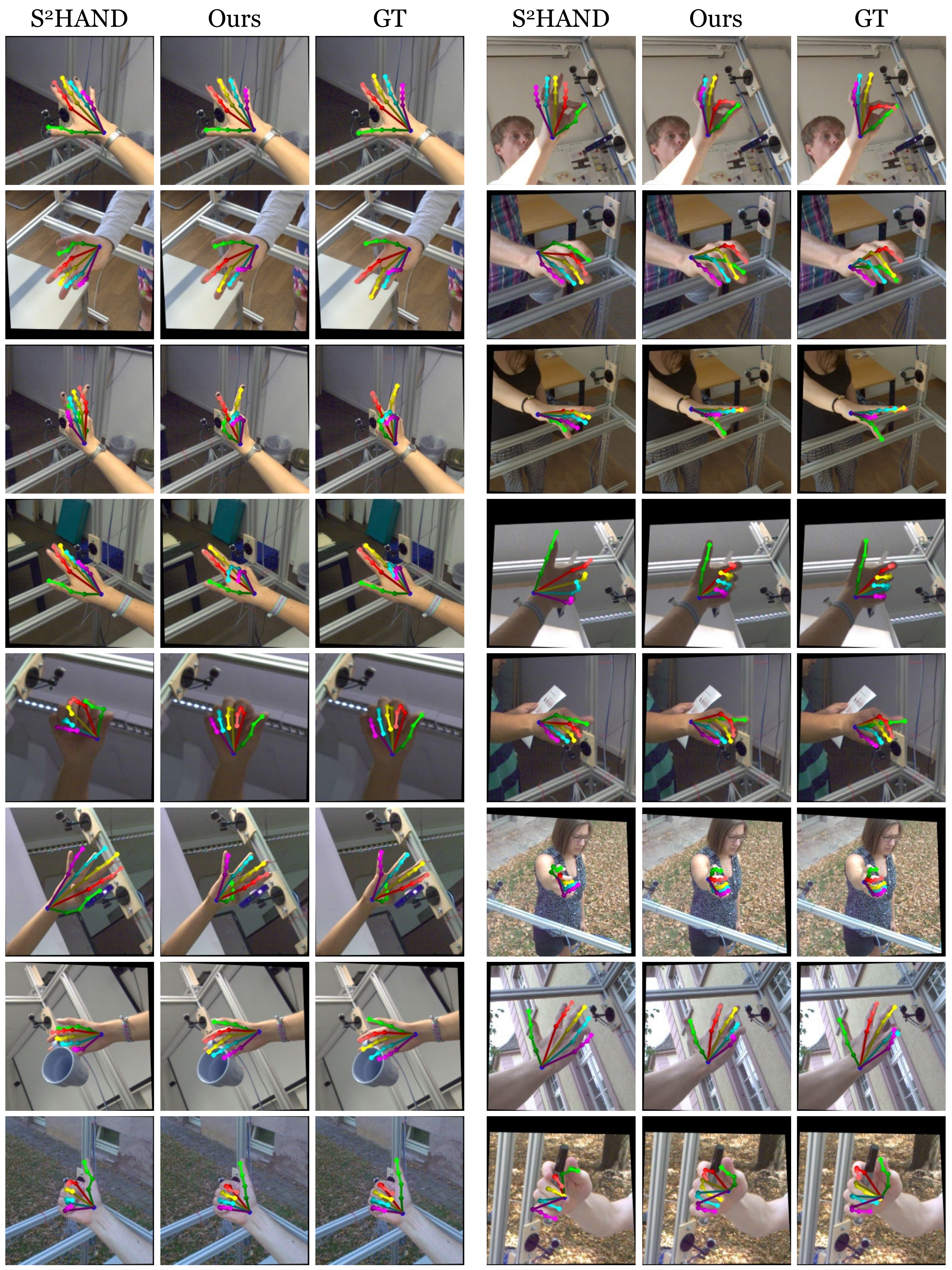} 
	\caption{2D prediction (overlayed in the images) comparisons between S$^{2}$HAND, ours, and the ground-truth on the FreiHAND dataset.}
	\label{img:freihand}
\end{figure*}

\begin{figure*}[!t]
	\centering
	\includegraphics[width=0.97\textwidth]{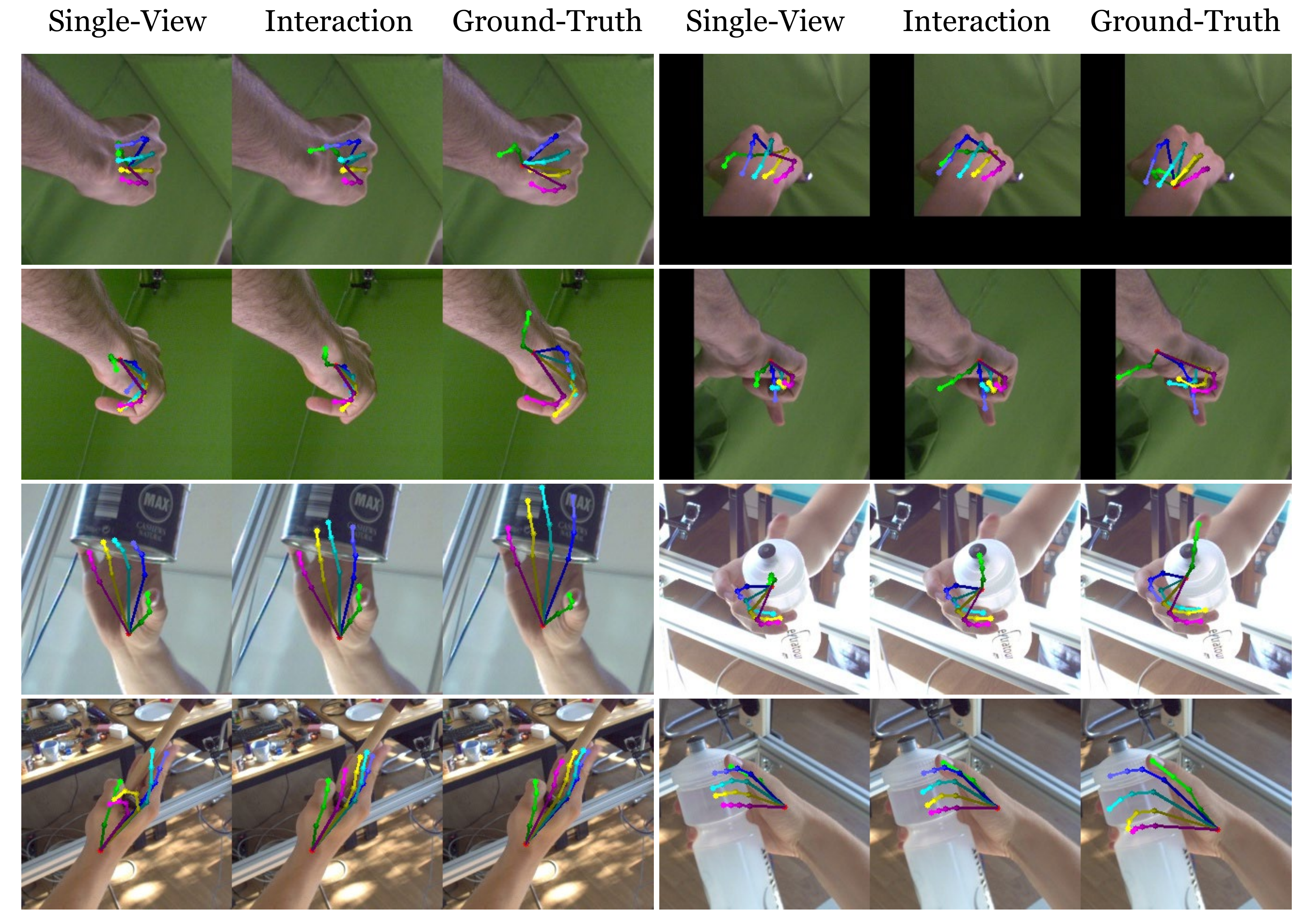} 
	\caption{2D prediction (overlayed in the images) of our failure cases on the HanCo dataset. From left to right, we show our predictions from the single-view network, cross-view interaction network, and the ground-truth.}
	\label{img:failure cases}
\end{figure*}

\clearpage
{\small
\bibliographystyle{ieee_fullname}
\bibliography{egbib}
}

\end{document}